%% file: main.tex
\documentclass{article}

\usepackage{amsmath,amsfonts,amssymb,amsthm,bm}
\usepackage{enumitem}
\usepackage{microtype}
\usepackage{graphbox}
\usepackage{graphicx}
\usepackage{caption}
\usepackage{subcaption}
\usepackage{wrapfig}
\usepackage{booktabs} 
\usepackage{amsfonts}       
\usepackage{nicefrac}       
\input{math_commands}
\usepackage{multirow}
\usepackage{makecell}
\usepackage{comment}

\usepackage{hyperref}

\usepackage[accepted]{icml2022}

\icmltitlerunning{Deep symbolic Regression for Recurrent Sequences}

\begin{document}

\twocolumn[
\icmltitle{Deep Symbolic Regression for Recurrent Sequences}



\icmlsetsymbol{equal}{*}

\begin{icmlauthorlist}
\icmlauthor{St\'ephane d'Ascoli}{equal,ens,fair}
\icmlauthor{Pierre-Alexandre Kamienny}{equal,fair,lip6}
\icmlauthor{Guillaume Lample}{fair}
\icmlauthor{François Charton}{fair}
\end{icmlauthorlist}

\icmlaffiliation{ens}{Department of Physics, \'Ecole Normale Sup\'erieure, Paris}
\icmlaffiliation{lip6}{Laboratoire d'Informatique de Paris 6, Sorbonne Université, Paris}
\icmlaffiliation{fair}{Meta AI, Paris}

\icmlcorrespondingauthor{Stéphane d'Ascoli}{stephane.dascoli@ens.fr}

\icmlkeywords{Machine Learning, ICML}

\vskip 0.3in
]



\printAffiliationsAndNotice{\icmlEqualContribution} 

\input{abstract}

\input{main_text}

\bibliography{refs}
\bibliographystyle{icml2022}

\clearpage
\appendix
\input{appendix}

\end{document}

%% file: math_commands.tex
\usepackage{xcolor}
\definecolor{C0}{HTML}{1f77b4}
\definecolor{C1}{HTML}{ff7f0e}
\definecolor{C2}{HTML}{2ca02c}
\definecolor{C3}{HTML}{d62728}
\definecolor{C4}{HTML}{9467bd}
\definecolor{C5}{HTML}{8c564b}
\definecolor{C6}{HTML}{e377c2}
\definecolor{C7}{HTML}{7f7f7f}
\definecolor{C8}{HTML}{bcbd22}
\definecolor{C9}{HTML}{17becf}

\usepackage[normalem]{ulem}

\newcommand{\arccosh}{\operatorname{arccosh}}
\newcommand{\arcsinh}{\operatorname{arcsinh}}
\newcommand{\arctanh}{\operatorname{arctanh}}

%% file: abstract.tex
\begin{abstract}
Symbolic regression, i.e. predicting a function from the observation of its values, is well-known to be a challenging task. In this paper, we train Transformers to infer the function or recurrence relation underlying sequences of integers or floats, a typical task in human IQ tests which has hardly been tackled in the machine learning literature. We evaluate our integer model on a subset of OEIS sequences, and show that it outperforms built-in Mathematica functions for recurrence prediction. We also demonstrate that our float model is able to yield informative approximations of out-of-vocabulary functions and constants, e.g. $\operatorname{bessel0}(x)\approx \frac{\sin(x)+\cos(x)}{\sqrt{\pi x}}$ and $1.644934\approx \pi^2/6$. 
\end{abstract}

%% file: main_text.tex
\section{Introduction} 
Given the sequence [1,2,4,7,11,16], what is the next term? Humans usually solve such riddles by noticing patterns in the sequence. In the easiest cases, one can spot a function: [1,4,9,16,25] are the first five squares, so the n-th term in the series is $u_n=n^2$, and the next one is 36. Most often however, we look for relations between successive terms: in the sequence [1,2,4,7,11,16], the differences between successive values are 1, 2, 3, 4, and 5, which makes it likely that the next term will be $16+6=22$. Mathematically, we are inferring the recurrence relation $u_{n} = u_{n-1} + n$, with $u_0=1$. 

In all cases, we handle such problems as symbolic regressions: starting from a sequence of numbers, we try to discover a function or a recurrence relation that they satisfy, and use it to predict the next terms. This can lead to very challenging problems as the complexity of the unknown recurrence relation $u_{n}=f(n, \{u_i\}_{i<n})$ increases, e.g. $u_{n} = \tan^{-1}(u_{n-3})\exp(\cos(n^2))$.
 
In this paper, we train neural networks to infer the recurrence relation $f$ from the observation of the first terms of the sequence. The majority of studies in machine learning for symbolic regression have focused on non-recurrent functions, i.e. expressions of the form $y=f(x)$. Recurrence relations provide a more general setting, which gives a deeper account of the underlying process: if the sequence corresponds to successive time steps, the recurrence relation is a discrete time differential equation for the system considered. 

\begingroup
\renewcommand{\arraystretch}{1.3} 
\begin{table*}[htb]
    \centering
    \small
    \begin{tabular}{c|c|c|c}
    \toprule
        OEIS & Description & First terms & Predicted recurrence \\
    \midrule
        A000792 & $a(n) = \max\{(n - i)a(i), i < n\}$ & 	1, 1, 2, 3, 4, 6, 9, 12, 18, 27 & $u_{n}= u_{n-1} + u_{n-3} - u_{n-1}\% u_{n-3}$ \\
        A000855 & Final two digits of $2^n$                         & 1, 2, 4, 8, 16, 32, 64, 28, 56, 12        & $u_{n}=(2 u_{n-1}) \% 100$ \\
        A006257 & Josephus sequence & 	0, 1, 1, 3, 1, 3, 5, 7, 1, 3 & $u_n = (u_{n-1}+n) \% (n - 1) - 1$ \\
        A008954 & Final digit of $n(n+1)/2$        & 0, 1, 3, 6, 0, 5, 1, 8, 6, 5              & $u_{n}=(u_{n-1} + n)\%10$\\
        A026741 & 	$a(n) = n$ if $n$ odd, $n/2$ if $n$ even  & 0, 1, 1, 3, 2, 5, 3, 7, 4, 9 & $u_{n}= u_{n-2} + n//(u_{n-1} + 1)$ \\
        A035327 & $n$ binary, switch 0's and 1's, then decimal  & 1, 0, 1, 0, 3, 2, 1, 0, 7, 6              & $u_{n}= (u_{n-1} - n) \% (n - 1)$ \\
        A062050 & $n$-th chunk contains numbers $1, ..., 2^n$  & 1, 1, 2, 1, 2, 3, 4, 1, 2, 3              & $u_{n}=(n \% (n - u_{n-1})) + 1$\\
        A074062 & Reflected Pentanacci numbers                      & 5, -1, -1, -1, -1, 9, -7, -1, -1, -1      & $u_{n}=2u_{n-5} - u_{n-6}$\\
    \bottomrule
    \end{tabular}
    \caption{\textbf{Our integer model yields exact recurrence relations on a variety of interesting OEIS sequences.} Predictions are based on observing the first 25 terms of each sequence.
    }
    \label{tab:oeis-examples}
\end{table*}
\endgroup

\begin{table}[htb]
    \centering
    \begin{tabular}{c|c|c}
    \toprule
        Constant & Approximation & Rel. error\\
        \midrule
        $0.3333      $ & $ (3+\exp(-6))^{-1}    $ & $10^{-5}$\\
        $0.33333     $ & $ 1/3                  $ & $10^{-5}$\\
        \hline
        $3.1415      $ & $ 2\arctan (\exp(10))     $ & $10^{-7}$\\
        $3.14159     $ & $ \pi                  $ & $10^{-7}$\\
        \hline
        $1.6449      $ & $ 1/\arctan (\exp(4))     $ & $10^{-7}$\\
        $1.64493     $ & $ \pi^2/6              $ & $10^{-7}$\\
        \hline
        $0.123456789 $ & $ 10/9^2               $ & $10^{-9}$\\
        $0.987654321 $ & $ 1-(1/9)^2            $ & $10^{-11}$\\
    \bottomrule
    \end{tabular}
    \caption{\textbf{Our float model learns to approximate out-of-vocabulary constants with its own vocabulary.} We obtain the approximation of each constant $C$ by feeding our model the 25 first terms of $u_n=Cn$.
    }
    \label{tab:approx-constants}
\end{table}

\begin{table}[htb]
    \centering
    \begin{tabular}{c|c|c}
    \toprule
        Expression $u_n$ & Approximation $\hat u_n$ & Comment\\
    \midrule
        $\arcsinh(n)$               & $\log(n + \sqrt{n^2 + 1})$                        & Exact\\
        $\arccosh(n)$               & $\log(n + \sqrt{n^2-1})$                          & Exact\\
        $\arctanh(1/n)$             & $\frac{1}{2}\log(1+2/n)$                          & Asymptotic\\
    \hline
        $\operatorname{catalan}(n)$  & $u_{n-1}(4-6/n)$                                 & Asymptotic\\
        $\operatorname{dawson}(n)$  & $\frac{n}{2n^2-u_{n-1}-1}$                        & Asymptotic\\
    \hline
        $\operatorname{j0}(n)$ (Bessel)  & $ \frac{\sin(n)+\cos(n)}{\sqrt{\pi n}} $     & Asymptotic\\
        $\operatorname{i0}(n)$ (mod. Bessel) & $ \frac{e^n}{\sqrt{2\pi n}} $            & Asymptotic\\
    \bottomrule
    \end{tabular}
    \caption{\textbf{Our float model learns to approximate out-of-vocabulary functions with its own vocabulary.} For simple functions, our model predicts an exact expression in terms of its operators; for complex functions which cannot be expressed, our model manages to predict the first order of the asymptotic expansion. 
    }
    \label{tab:approx-functions}
\end{table}

\subsection{Contributions}
We show that transformers can learn to infer a recurrence relation from the observation of the first terms of a sequence. We consider both sequences of integers and floats, and train our models on a large set of synthetic examples.

We first demonstrate that our symbolic regression model can predict complex recurrence relations that were not seen during training. We also show that those recurrence relations can be used to extrapolate the next terms of the sequence with better precision than a ``numeric'' model of similar architecture, trained specifically for extrapolation. 

We then test the out-of-domain generalization of our models. On a subset of the Online Encyclopedia of Integer Sequences, our integer models outperform built-in Mathematica functions, both for sequence extrapolation and recurrence prediction, see Table~\ref{tab:oeis-examples} for some examples. We also show that our symbolic float model is capable of predicting approximate expressions for out-of-vocabulary functions and constants (e.g. $\operatorname{bessel0}(x)\approx \frac{\sin(x)+\cos(x)}{\sqrt{\pi x}}$ and $1.644934\approx \pi^2/6$), see Tables~\ref{tab:approx-constants},~\ref{tab:approx-functions} for more examples. 

We conclude by discussing potential limitations of our models and future directions.

\paragraph{Demonstration} We provide an interactive demonstration of our models at~\url{https://symbolicregression.metademolab.com}. 

\paragraph{Code} An open-source implementation of our code will be released publicly at \url{https://github.com/facebookresearch/recur}.

\subsection{Related work}

\paragraph{AI for maths} The use of neural networks for mathematics has recently gained attention: several works have demonstrated their surprising reasoning abilities~\cite{saxton2019analysing,cobbe2021training}, and have even sparked some interesting mathematical discoveries~\cite{davies2021advancing}. In particular, four types of tasks have recently been tackled in the deep learning literature. 

First, converting a symbolic expression to another symbolic expression. Direct examples are integration and differentation~\cite{lample2019deep}, expression simplification~\cite{allamanis2017learning} or equation solving~\cite{arabshahi2018towards}. Second, converting a symbolic expression to numerical data. This includes predicting quantitative properties of a mathematical structure, for example the local stability of a differential system~\cite{charton2020learning}. Third, converting numerical data to numerical data using mathematical rules. Examples range from learning basic arithmetics~\cite{kaiser2015neural,trask2018neural} to linear algebra~\cite{charton2021linear}. Fourth, converting numerical data to a symbolic expression: this is the framework of symbolic regression, which we will be focusing on. 

\paragraph{Symbolic regression} Two types of approaches exist for symbolic regression, which one could name selection-based and pretraining-based. 

In selection-based approaches, we only have access to one set of observations: the values of the function we are trying to infer. Typical examples of this approach are evolutionnary programs, which have long been the \textit{de facto} standard for symbolic regression~\cite{augusto2000symbolic,schmidt2009distilling,murari2014symbolic,mckay1995using}, despite their computational cost and limited performance. More recently, neural networks were used following this approach~\cite{sahoo2018learning,kim2020integration,petersen2019deep}. 

In pretraining-based approaches, we train neural networks on a large dataset containing observations from many different function~\cite{biggio2021neural,valipour2021symbolicgpt}, hoping that the model can generalize to new expressions. Although the pretraining is computationally expensive, inference is much quicker as one simply needs to perform a forward pass, rather than search through a set of functions. In this work, we choose this approach. 

\paragraph{Recurrent sequences} All studies on symbolic regression cited above consider the task of learning non-recurrent functions from their values at a set of points. Our contribution is, to the best of our knowledge, the first to tackle the setup of recurrent expressions. Although this includes non-recurrent expressions as a special case, one slight restriction is that the inputs need to be sampled uniformly. Hence, instead of feeding the model with input-output pairs $\{(x_i,y_i)\}$, we only need to feed it the terms of the sequence $\{u_i\}$. Another important difference is that order of these terms matters, hence permutation invariant representations~\cite{valipour2021symbolicgpt} cannot be used.

Integer sequences, in particular those from the Online Encyclopedia of Integer Sequences (OEIS)~\cite{sloane2007line}, have been studied using machine learning methods in a few recent papers. \citet{wu2018can} trains various classifiers to predict the label of OEIS sequences, such as ``interesting'', ``easy'', or ``multiplicative''. \citet{ryskina2021learning} use embeddings trained on OEIS to investigate the mathematical properties of integers. \citet{ragni2011} use fully connected networks to predict the next term of OEIS sequences (a numeric rather than symbolic regression task). \citet{nam2018number} investigate different architectures (most of them recurrent networks) for digit-level numeric regression on sequences such as Fibonacci, demonstrating the difficulty of the task. 

\section{Methods}

Broadly speaking, we want to solve the following problem: given a sequence of $n$ points $\{u_0, \ldots, u_{n-1}\}$, find a function $f$ such that for any $i\in \mathbb{N}$, $u_i = f(i, u_{i-1}, \ldots, u_{i-d})$, where $d$ is the recursion degree. Since we cannot evaluate at an infinite number of points, we declare that a function $f$ is a solution of the problem if, given the first $n_{input}$ terms in the sequence, it can predict the next $n_{pred}$ following ones.

Under this form, the problem is underdetermined: given a finite number of input terms, an infinite number of recurrence relations exist. However in practice, one would like the model to give us the simplest solution possible, following Occam's principle. This is generally ensured by the fact that simple expressions are more likely to be generated during training.

\paragraph{Integer vs. float} We consider two settings for the numbers in the sequences: integers and floats. For integer sequences, the recurrence formula only uses operators which are closed in $\mathbb{Z}$ (e.g. $+, \times, \mathrm{abs}$, modulo and integer division). For float sequences, additional operators and functions are allowed, such as real division, $\mathrm{exp}$ and $\mathrm{cos}$ (see Table~\ref{tab:operators} for the list of all used operators). These two setups are interesting for different reasons. Integer sequences are an area of strong interest in mathematics, particular for their relation with arithmetics. The float setup is interesting to see how our model can generalize to a larger set of operators, which provides more challenging problems.

\paragraph{Symbolic vs. numeric} We consider two tasks: symbolic regression and numeric regression. In symbolic regression, the model is tasked to predict the recurrence relation the sequence was generated from. At test time, this recurrence relation is evaluated by how well it approximates the $n_{pred}$ following terms in the sequence. 

In numeric regression, is tasked to directly predict the values of the $n_{pred}$ following terms, rather than the underlying recurrence relation. At test time, the model predictions are compared with the true values of the sequence. 


\begin{table}[htb]
    \centering
    \small
    \begin{tabular}{c|c|c}
    \toprule
        & Integer & Float\\
        \midrule
        Unary & \makecell{\texttt{abs, sqr,}\\\texttt{sign, relu}} & \makecell{\texttt{abs, sqr, sqrt,}\\\texttt{inv, log, exp,}\\\texttt{sin, cos, tan, atan}}\\
        \hline
        Binary & \makecell{\texttt{add, sub, mul,}\\\texttt{intdiv, mod}} & \texttt{add, sub, mul, div}\\
    \bottomrule
    \end{tabular}
    \caption{\textbf{Operators used in our generators.}}
    \label{tab:operators}
\end{table}

\subsection{Data generation}
\label{sec:data-generation}

All training examples are created by randomly generating a recurrence relation, randomly sampling its initial terms, then computing the next terms using the recurrence relation.  More specifically, we use the steps below:

\begin{enumerate}
    \item Sample the \textbf{number of operators} $o$ between $1$ and $o_{max}$, and build a unary-binary tree with $o$ nodes, as described in~\cite{lample2019deep}. The number of operators determines the difficulty of the expression.
    \item Sample the \textbf{nodes} of the tree from the list of operators in Table~\ref{tab:operators}. Note that the float case uses more operators than the integer case, which makes the task more challenging by expanding the problem space.
    \item Sample the \textbf{recurrence degree} $d$ between 1 and $d_{max}$, which defines the recurrence depth: for example, a degree of $2$ means that $u_{n+1}$ depends on $u_{n}$ and $u_{n-1}$.
    \item Sample the \textbf{leaves} of the tree: either a constant, with probability $p_{const}$, or the current index $n$, with probability $p_n$, or one of the previous terms of the sequence $u_{n-i}$, with $i\in [1,d]$, with probability $p_{var}$. 
    \item Recalculate the true recurrence degree $d$ considering the deepest leaf $u_{n-i}$ sampled during the previous step, then sample $d$ \textbf{initial terms} from a random distribution $\mathcal{P}$.
    \item Sample $l$ between $l_{min}$ and $l_{max}$ and compute the next $l$ terms of the sequence using the initial conditions. The total \textbf{sequence length} is hence $n_{input}=d_\mathrm{eff}+l$.
\end{enumerate}

We provide the values of the parameters of the generator in Table~\ref{tab:generator}. Note that in the last step, we interrupt the computation if we encounter a term larger than $10^{100}$, or outside the range of one of the operators: for example, a division by zero, or a negative square root. 

\begin{table}[htb]
    \centering
    \begin{tabular}{c|c|c}
    \toprule
    Parameter & Description & Value\\
    \midrule
        $d_{max}$ & Max degree                     & $6        $\\
        $o_{max}$ & Max number of operators              & $10       $\\
        $l_{min}$ & Min length                     & $5        $\\
        $l_{max}$ & Max length                     & $30       $\\
        $p_{const}$ & Prob of constant leaf        & $1/3      $\\
        $p_{index}$ & Prob of index leaf           & $1/3      $\\
        $p_{var}$   & Prob of variable leaf        & $1/3      $\\
        $\mathcal{P}$ & Distrib of first terms     & $\mathcal{U}(-10,10)$\\
    \bottomrule
    \end{tabular}
    \caption{\textbf{Hyperparameters of our generator.}}
    \label{tab:generator}
\end{table}

\paragraph{Limitations}

One drawback of our approach is poor out-of-distribution generalization of deep neural networks, demonstrated in a similar context to ours by~\cite{charton2020learning}: when shown an example which is impossible to be generated by the procedure described above, the model will inevitably fail. For example, as shown in App.~\ref{app:ood-init}, our model performs poorly when the initial terms of the sequence are sampled outside their usual range, which is chosen to be $[-10, 10]$ in these experiments. One easy fix is to increase this range; are more involved one is to use a normalization procedure as described in our follow-up work~\cite{kamienny2022end}.

\subsection{Encodings}
\label{sec:encoding}

Model inputs are sequences of integers or floats. The outputs are recurrence relations for the symbolic task, and sequences of numbers for the numeric task. To be processed by transformers, inputs and outputs must be represented as sequences of tokens from a fixed vocabulary. To this effect, we use the encodings presented below.

\paragraph{Integers} We encode integers using their base $b$ representation, as a sequence of $1 + \lceil\log_{b} |x|\rceil$ tokens: a sign (which also serves as a sequence delimiter) followed by $\lceil\log_{b} |x|\rceil$ digits from $0$ to $b-1$, for a total vocabulary of $b+2$ tokens. For instance, $x=-325$ is represented in base $b=10$ as the sequence \texttt{[-, 3, 2, 5]}, and in base $b=30$ as \texttt{[-, 10, 25]}. 
Choosing $b$ involves a tradeoff between the length of the sequences fed to the Transformer and the vocabulary size of the embedding layer. We choose $b=10,000$ and limit integers in our sequences to absolute values below $10^{100}$, for a maximum of $26$ tokens per integer, and a vocabulary of order $10^4$ words.

\paragraph{Floats} Following~\citet{charton2021linear}, we represent float numbers in base $10$ floating-point notation, round them to four significant digits, and encode them as sequences of $3$ tokens: their sign, mantissa (between \texttt{0} and \texttt{9999}), and exponent (from \texttt{E-100} to \texttt{E100}). For instance, $1/3$ is encoded as $\texttt{[+, 3333, E-4]}$. Again, the vocabulary is of order $10^4$ words.

For all operations in floating-point representation, precision is limited to the length of the mantissa. In particular, when summing elements with different magnitudes, sub-dominant terms may be rounded away. Partly due to this effect, when approximating complex functions, our symbolic model typically only predicts the largest terms in its asymptotic expansion, as shown in Table~\ref{tab:approx-functions}. We discuss two methods for increasing precision when needed in Section~\ref{app:iterative-refinement} of the Appendix.

\paragraph{Recurrence relations}
To represent mathematical trees as sequences, we enumerate the trees in prefix order, i.e. direct Polish notation, and encode each node as a single autonomous token. For instance, the expression $\cos(3x)$ is encoded as $\texttt{[cos,mul,3,x]}$.

Note that the generation procedure implies that the recurrence relation is not simplified (i.e. expressions like $1+u_{n-1}-1$ can be generated). We tried simplifying them using Sympy before the encoding step (see Appendix~\ref{app:sympy}), but this slows down generation without any benefit on the performance of our models, which turn out to be surprisingly insensitive to the syntax of the formulas.

\subsection{Experimental details}

Similarly to~\citet{lample2019deep}, we use a simple Transformer architecture~\cite{vaswani2017attention} with 8 hidden layers, 8 attention heads and an embedding dimension of 512 both for the encoder and decoder.

\paragraph{Training and evaluation}
The tokens generated by the model are supervised via a cross-entropy loss. We use the Adam optimizer, warming up the learning rate from $10^{-7}$ to $2.10^{-4}$ over the first 10,000 steps, then decaying it as the inverse square root of the number of steps, following \citep{vaswani2017attention}.
We train each model for a minimum of 250 epochs, each epoch containing 5M equations in batches of 512. On 16 GPU with Volta architecture and 32GB memory, one epoch is processed in about an hour.

After each epoch, we evaluate the in-distribution performance of our models on a held-out dataset of 10,000 equations. Unless specified otherwise, we generate expressions using greedy decoding. Note that nothing prevents the symbolic model from generating an invalid expression such as \texttt{[add,1,mul,2]}. These mistakes, counted as invalid predictions, tend to be very rare: in models trained for more than a few epochs, they occur in less than 0.1\% of the test cases.

\paragraph{Hypothesis ranking}
To predict recurrence relations for OEIS sequences (Section~\ref{sec:oeis}), or the examples displayed in Tables~\ref{tab:oeis-examples},\ref{tab:approx-constants},\ref{tab:approx-functions}, we used a beam size of 10. Usually, hypotheses in the beam are ranked according to their log-likelihood, normalized by the sequence length. For our symbolic model, we can do much better, by ranking beam hypotheses according to how well they approximate the initial terms of the original sequence. Specifically, given a recurrence relation of degree $d$, we recompute for each hypothesis the $n_{input}$ first terms in the sequence (using the first $d$ terms of the original sequence as initial conditions), and rank the candidates according to how well they approximate these terms. This ranking procedure is impossible for the numerical model, which can only perform extrapolation. 

In some sense, this relates our method to evolutionary algorithms, which use the input terms for candidate selection. Yet, as shown by the failure modes presented in Section~\ref{app:examples} of the Appendix, our model is less prone to overfitting since the candidates are not directly chosen by minimizing a loss on the input terms.

\section{In-domain generalization}

We begin by probing the in-domain accuracy of our model, i.e. its ability to generalize to unseen sequences generated with the same procedure as the training data. As discussed in Section~\ref{app:memorization} of the Appendix, the diversity of mathematical expressions and the random sampling of the initial terms ensures that almost all the examples presented at test time have not been seen during training: one cannot attribute the generalization to mere memorization.

\paragraph{Prediction accuracy}

Due to the large number of equivalent ways one can represent a mathematical expression, one cannot directly evaluate the accuracy by comparing (token by token) the predicted recurrence relation to the ground truth. Instead, we use the predicted expression to compute the next $n_{pred}$ terms of the sequence $\{\hat u_i\}$, and compare them with those computed from the ground truth, $\{u_i\}$. The prediction accuracy is then defined as:
\begin{align}
    acc(n_{pred},\tau) = \mathbb{P}\left(\max_{1\leq i \leq n_{pred}} \left\vert \frac{\hat u_i - u_i}{u_i}\right\vert \leq \tau \right)
\end{align}

\begin{figure*}
    \centering
    \includegraphics[width=\linewidth]{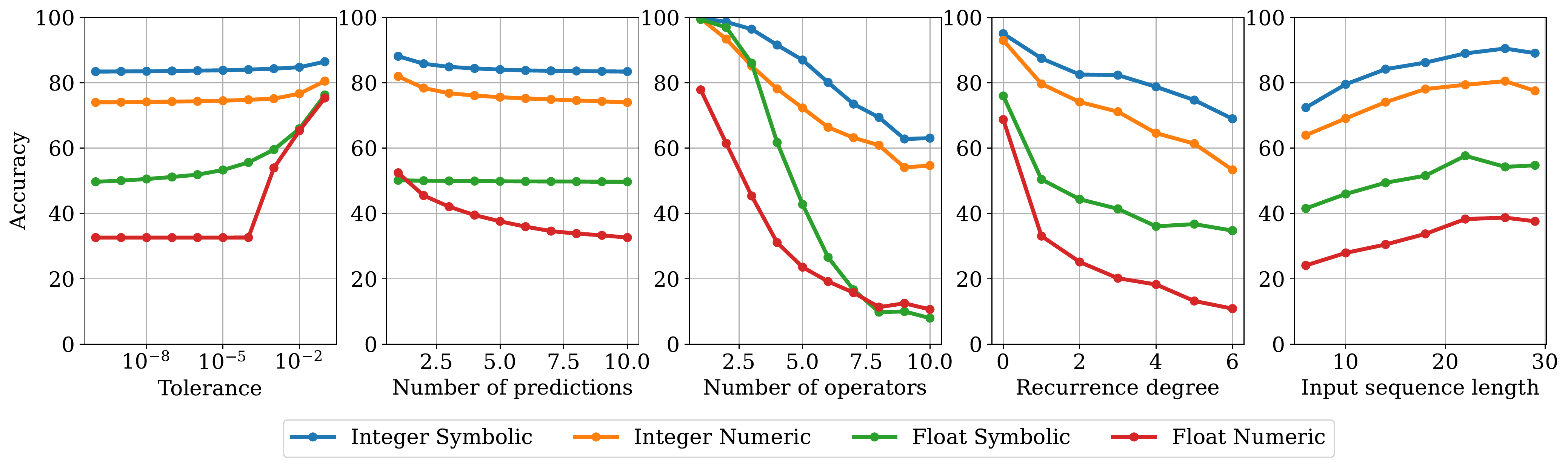}
    \caption{\textbf{The symbolic model extrapolates further and with higher precision than the numeric model.} From left to right, we vary the tolerance $\tau$, the number of predictions $n_{pred}$, the number of operators $o$, the recurrence degree $d$ and the number of input terms $l$. In each plot, we use the following defaults for quantities which are not varied: $\tau=10^{-10}$, $n_{pred}=10$, $o\in[\![1,10]\!]$, $d\in[\![1,6]\!]$, $l\in[\![5,30]\!]$.}
    \label{fig:acc-difficulty}
\end{figure*}

By choosing a small enough $\tau \geq 0$ and a large enough $n_{pred}$, one can ensure that the predicted formula matches the true formula. In the float setup, $\tau=0$ must be avoided for two reasons: (i) equivalent solutions represented by different expressions may evaluate differently because due to finite machine precision, (ii) setting $\tau=0$ would penalize the model for ignoring sub-dominant terms which are indetectable due to the finite precision encodings. Hence, we select $\tau=10^{-10}$ and $n_{pred}=10$ unless specified otherwise\footnote{For the float numeric model, which can only predict values up to finite precision $\epsilon$, we round the values of target function to the same precision. This explains the plateau of the accuracy at $\tau<\epsilon$ in Figure~\ref{fig:acc-difficulty}.}. Occasionally, we will consider larger values of $\tau$, to assess the ability of our model to provide approximate expressions. 

\begin{table}[tb]
    \centering
    \begin{tabular}{c|cc|cc}
    \toprule
        \multirow{2}{*}{Model} & \multicolumn{2}{c|}{Integer} & \multicolumn{2}{c}{Float}\\
        & $n_{op}\!\leq\!5$ & $n_{op}\!\leq\!10$ & $n_{op}\!\leq\!5$ & $n_{op}\!\leq\!10$ \\
        \midrule
        Symbolic & \textbf{92.7} & \textbf{78.4} & \textbf{74.2} & \textbf{43.3} \\
        Numeric  & 83.6 & 70.3 & 45.6 & 29.0 \\
    \bottomrule
    \end{tabular}
    \caption{\textbf{Average in-distribution accuracies of our models.} We set $\tau=10^{-10}$ and $n_{pred}=10$.}
    \label{tab:acc}
\end{table}

\paragraph{Results}

The average in-distribution accuracies of our models are reported in Table~\ref{tab:acc} with $\tau=10^{-10}$ and $n_{pred}=10$. Although the float setup is significantly harder that the integer setup, our symbolic model reaches good accuracies in both cases. In comparison, the numeric model obtains worse results, particularly in the float setup. The interested reader may find additional training curves, attention maps and visualizations of the model predictions in Appendix~\ref{app:examples}.

\paragraph{Ablation over the evaluation metric}

The two first panels of Figure~\ref{fig:acc-difficulty} illustrate how the accuracy changes as we vary the tolerance level $\tau$ and the number of predictions $n_{pred}$.
The first important observation is that the symbolic model performs much better than the numeric model at low tolerance. At high tolerance, it still keeps an advantage in the integer setup, and performs similarly in the float setup. This demonstrates the advantage of a symbolic approach for high-precision predictions.

Second, we see that the accuracy of both models degrades as we increase the number of predictions $n_{pred}$, as one could expect. However, the decrease is less important for the symbolic model, especially in the float setup where the curve is essentially flat. This demonstrates another strong advantage of the symbolic approach: once it has found the correct formula, it can predict the whole sequence, whereas the precision of the numeric model deteriorates as it extrapolates further. 

\paragraph{Ablation over the example difficulty}

The three last panels of Figure~\ref{fig:acc-difficulty} decompose the accuracy of our two models along three factors of  difficulty: the number of operators $o$\footnote{Since expressions are not simplified, $o$ may be overestimated.}, the recurrence degree $d$ and the sequence length $n_{input}$ (see Section~\ref{sec:data-generation}).  

Unsurprisingly, accuracy degrades rapidly as the number of operators increases, particularly in the float setting where the operators are more diverse: the accuracy of the symbolic model drops from 100\% for $o\!=\!1$ to $10\%$ for $o\!=\!10$. We attempted a curriculum learning strategy to alleviate the drop, by giving higher probabilities to expressions with many operators as training advances, but this did not bring any improvement. Increasing the recurrence degree has a similar but more moderate effect: the accuracy decreases from 70\% for $d\!=\!0$ (non-recurrent expressions) to 20\% for $d\!=\!6$ in the float setup. Finally, we observe that shorter sequences are harder to predict as they give less information on the underlying recurrence; however, even when fed with less than 10 terms, our models achieve surprisingly high accuracies.

\begin{figure}[tb]
    \centering
    \begin{subfigure}[b]{.49\linewidth}
    \includegraphics[width=\linewidth]{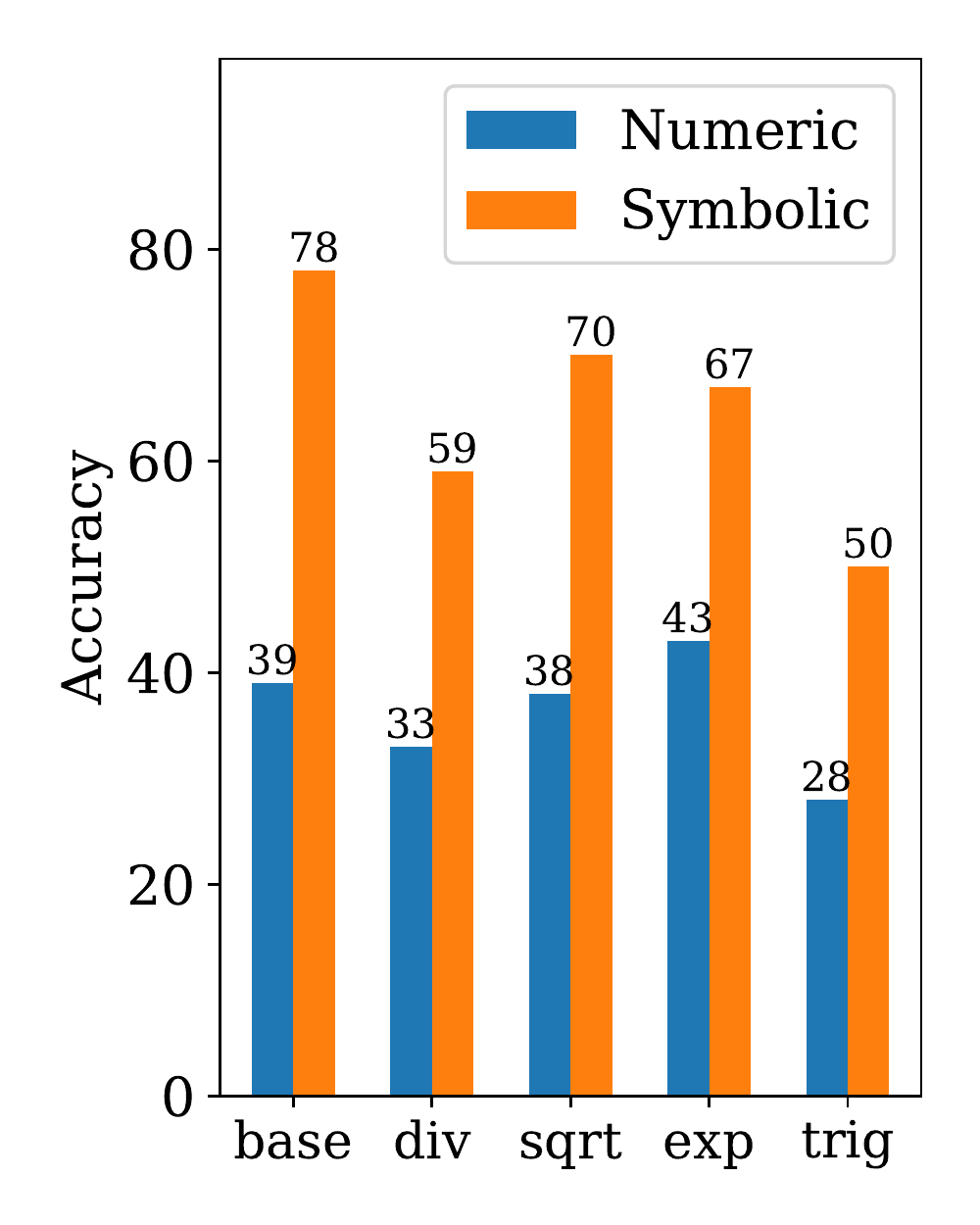}
    \caption{In-domain}
    \end{subfigure}
    \begin{subfigure}[b]{.49\linewidth}
    \includegraphics[width=\linewidth]{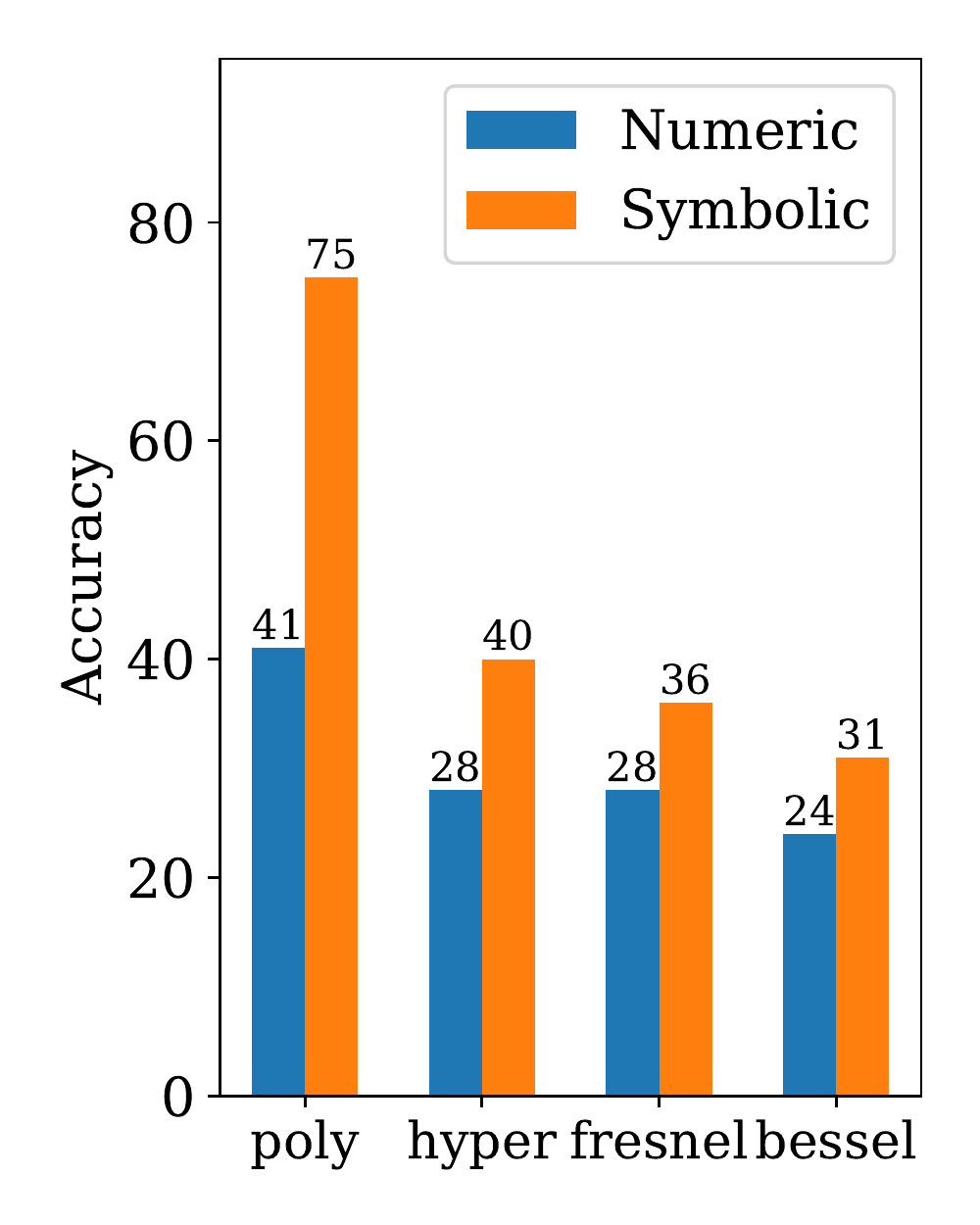}
    \caption{Out-of-domain}
    \end{subfigure}
    \caption{\textbf{Accuracy of our models on various in-domain and out-of-domain groups. } We set $\tau=10^{-10}$, $n_{pred}=10$. 
    }
    \label{fig:operator-families}
\end{figure}

\paragraph{Ablation over operator families}

To understand what kind of operators are the hardest for our float model, we bunch them into 5 groups:
\begin{itemize}[noitemsep]
    \item \textbf{base}:     \{add, sub, mul\}.
    \item \textbf{division}:       base + \{div, inv\}.
    \item \textbf{sqrt}: base + \{sqrt\}.
    \item \textbf{exponential}:    base + \{exp, log\}.
    \item \textbf{trigonometric}:   base + \{sin, cos, tan, arcsin, arccos, arctan\}.
\end{itemize}

Results are displayed in Figure~\ref{fig:operator-families}a. We see that the main difficulties lie in division and trigonometric operators, but the performance of both models stays rather good in all categories.

\paragraph{Visualizing the embeddings}

\begin{figure}[htb]
    \centering
    \begin{subfigure}[b]{.48\columnwidth}
    \includegraphics[width=\linewidth]{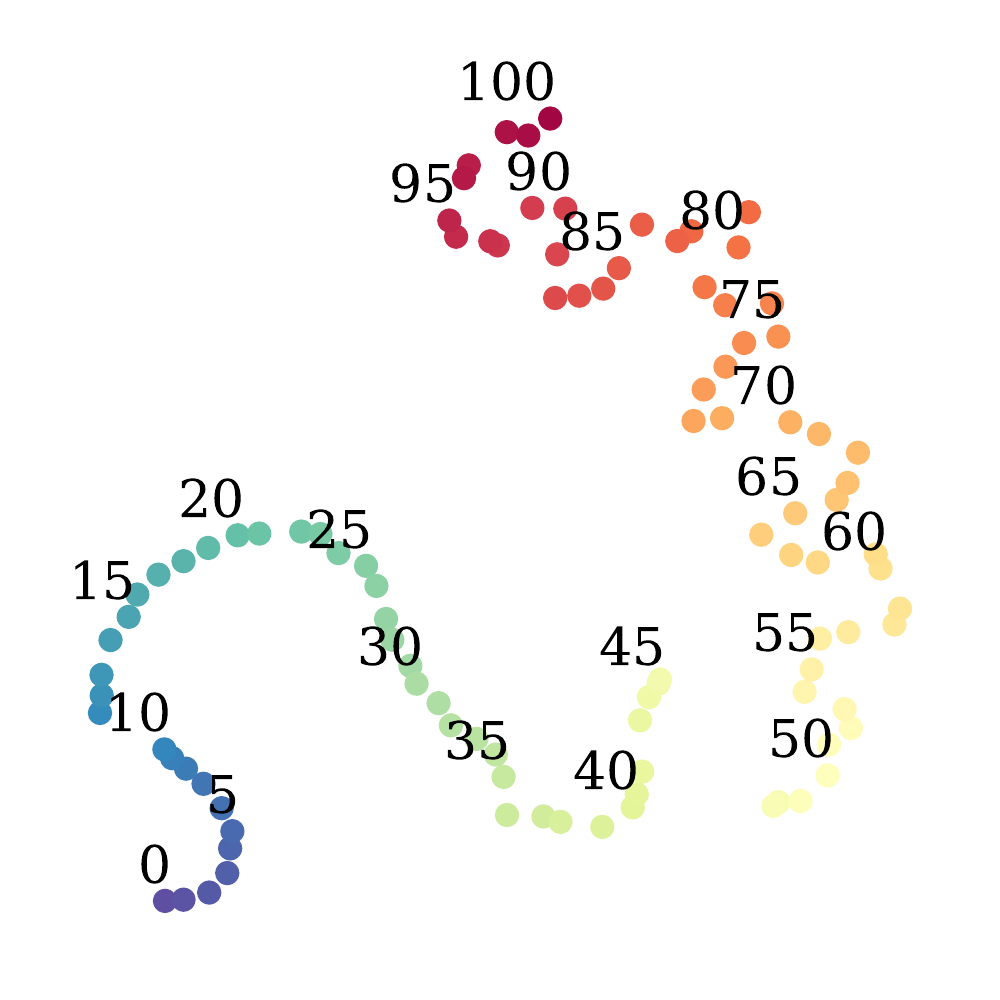}
    \caption{Integer embeddings}
    \end{subfigure} 
    \begin{subfigure}[b]{.48\columnwidth}
    \includegraphics[width=\linewidth]{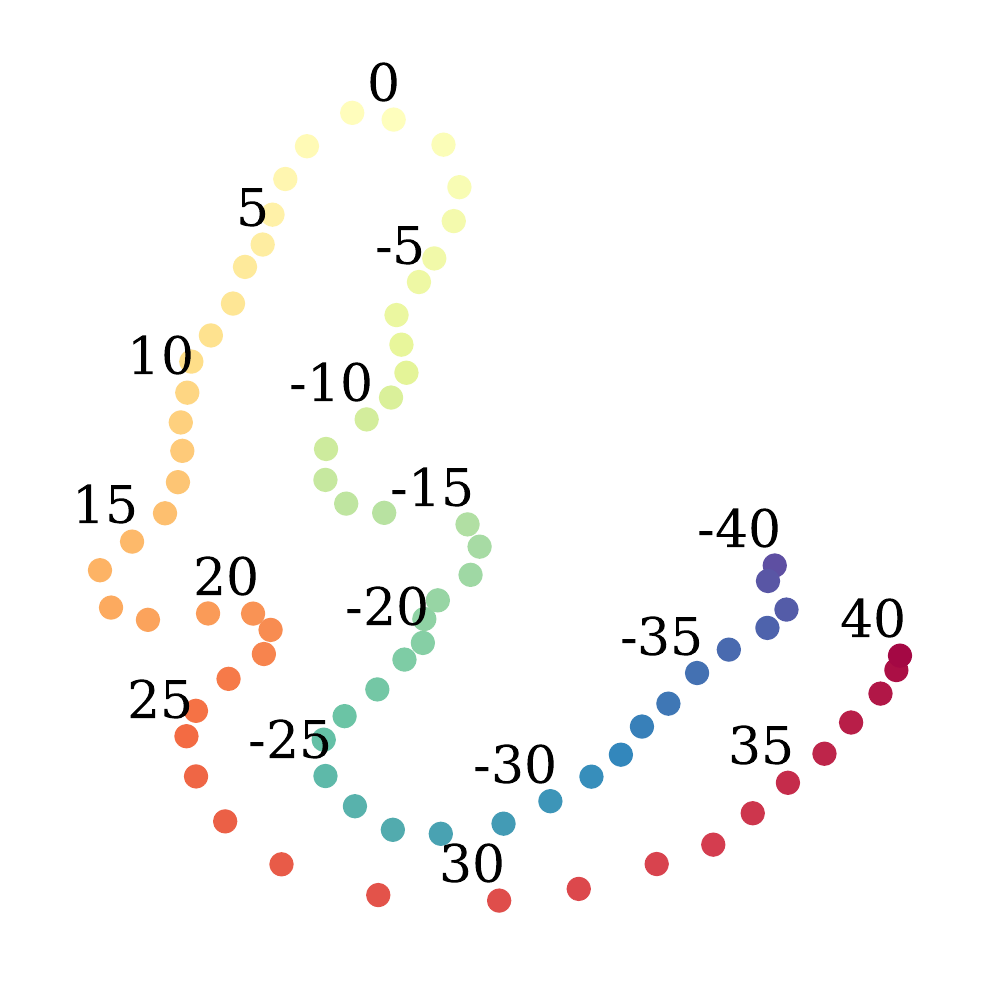}
    \caption{Exponent embeddings}
    \end{subfigure}
    \caption{\textbf{The number embeddings reveal intriguing mathematical structure}. We represented the t-SNE of the embeddings of the integer model and the exponent embeddings of the float model. We depicted the first 100 integer embeddings (10,000 in the model), and the exponent embeddings -40 to 40 (-100 to 100 in the model).}
    \label{fig:embeddings}
\end{figure}

To give more intuition on the inner workings of our symbolic models, we display a t-SNE \citep{vanDerMaaten2008} projection of the embeddings of the integer model in Figure~\ref{fig:embeddings}a and of exponent embeddings of the float model in Figure~\ref{fig:embeddings}b. 

Both reveal a sequential structure, with the embeddings organized in a clear order, as highlighted by the color scheme. In Appendix~\ref{app:embeddings}, we study in detail the pairwise distances between embeddings, unveiling interesting features such as the fact that the integer model naturally learns a base-6 representation.

\section{Out-of-domain generalization}

In this section, we evaluate the ability of our model to generalize out-of-domain. Recurrence prediction being a previously unexplored branch of symbolic regression, there are no official benchmarks we can compare our models to. For integer sequences, we use a subset of OEIS as our out-of-domain benchmark; for float sequences, we use a generator with out-of-vocabulary constants and operators. In Appendix~\ref{app:stochastic}, we also show that our models can and can be made robust noise in the inputs.

\subsection{Integer sequences: OEIS dataset}
\label{sec:oeis}

The Online Encyclopedia of Integer Sequences (OEIS) is an online database containing over 300,000 integer sequences. It is tempting to directly use OEIS as a testbed for prediction; however, many sequences in OEIS do not have a closed-form recurrence relation, such as the stops on the New York City Broadway line subway (A000053). These will naturally cause our model to fail.

\paragraph{Preprocessing} Luckily, OEIS comes with keywords, and 22\% of the sequences are labelled as ``easy'', meaning that there is a logic to find the next terms (although this logic is by no means easy in most cases). Note that this logic cannot always be formulated as a recurrence relation: for example, the sequence of primes or decimals of $\pi$ are included in this category, but intractable for our models. We keep the first 10,000 of these sequences as our testbed. Evaluation consists in in showing our models the first $n_{input}\in\{15,25\}$ terms of each sequence and asking it to predict the $n_{pred}\in\{1,10\}$ following terms. 

\paragraph{Results} Results are reported in Table~\ref{tab:oeis}. With only $n_{input}=15$ terms, the numeric model reaches an impressive accuracy of 53\% at next term prediction, and 27\% for predicting the next ten terms. The symbolic model achieves lower results, with 33\% and 19\% respectively; we attribute this to the large number of non-analytic sequences in the testbed. Nonetheless, this means that our model can retrieve a valid recurrence relation for almost a fifth of the sequences, which is rather impressive: we give a few interesting examples in Table~\ref{tab:oeis-examples}. Increasing $n_{input}$ to 25 increases our performances rather marginally.

As a comparison, we ran two built-in Mathematica functions for the task at hand: FindSequenceFunction, which finds non-recurrent expressions, and FindLinearRecurrence, which finds linear recurrence relations. These functions are much more sensitive to the number of terms given as input: they obtain similar accuracies at $n_{input}=15$, but FindLinearRecurrence performs significantly better at $n_{input}=25$, while FindSequenceFunction performs pathologically worse. Both these functions perform less well than our symbolic model in all cases.  

\begin{table*}[htb]
    \centering
    \begin{tabular}{c|cc|cc}
    \toprule
        \multirow{2}{*}{Model} & \multicolumn{2}{c|}{$n_{input}=15$} & \multicolumn{2}{c}{$n_{input}=25$}\\
        & $n_{pred}=1$ & $n_{pred}=10$ & $n_{pred}=1$ & $n_{pred}=10$ \\
        \midrule
        Symbolic (ours)         & 33.4 & 19.2 & 34.5 & 21.3 \\
        Numeric (ours)          & 53.1 & 27.4 & 54.9 & 29.5 \\
        FindSequenceFunction    & 17.1 & 12.0 & 8.1  & 7.2  \\ 
        FindLinearRecurrence    & 17.4 & 14.8 & 21.2 & 19.5 \\
    \bottomrule
    \end{tabular}
    \caption{\textbf{Accuracy of our integer models and Mathematica functions on OEIS sequences.} We use as input the first $n_{input}=\{15,25\}$ first terms of OEIS sequences and ask each model to predict the next $n_{pred}=\{1,10\}$ terms. We set the tolerance $\tau=10^{-10}$. }
    \label{tab:oeis}
\end{table*}

\subsection{Float sequences: robustness to out-of-vocabulary tokens}

One major difficulty in symbolic mathematics is dealing with out-of-vocabulary constants and operators: the model is forced to approximate them using its own vocabulary. We investigate these two scenarios separately for the float model.

\begin{table}[htb]
    \centering
    \small
    \begin{tabular}{c|cc|cc}
    \toprule
        \multirow{2}{*}{Model} & \multicolumn{2}{c|}{$[\![-10, 10]\!]\cup \{e,\pi,\gamma\}$} & \multicolumn{2}{c}{$\mathcal{U}(-10,10)$}\\
        & $n_{op}\!\leq\!5$ & $n_{op}\!\leq\!10$ & $n_{op}\!\leq\!5$ & $n_{op}\!\leq\!10$ \\
        \midrule
        Symbolic & \textbf{81.9} & \textbf{60.7} & 60.1 & 42.1 \\
        Numeric  & 72.4 & 60.4 & \textbf{72.2} & \textbf{60.2} \\
    \bottomrule
    \end{tabular}
    \caption{\textbf{Our symbolic model can approximate out-of-vocabulary prefactors.} We report the accuracies achieved when sampling the constants uniformly from $[\![-10, 10]\!] \cup \{e,\pi,\gamma\}$, as during training, versus sampling uniformly in $[-10,10]$. We set $\tau=0.01$ (note the higher tolerance threshold as we are considering approximation) and $n_{pred}=10$.}
    \label{tab:constants-perfs}
\end{table}

\paragraph{Out-of-vocabulary constants}

The first possible source of out-of-vocabulary tokens are prefactors. For example, a formula as simple as $u_n = 0.33 n$ is hard to predict perfectly, because our decoder only has access to integers between $-10$ and $10$ and a few mathematical constants, and needs to write $0.33$ as $[\texttt{div,add,mul,3,10,3,mul,10,10}]$. To circumvent this issue, ~\citet{biggio2021neural} use a separate optimizer to fit the prefactors, once the skeleton of the equation is predicted. 

In contrast, our model is end-to-end, and is surprisingly good at approximating out-of-vocabulary prefactors with its own vocabulary. For $u_n = 0.33n$, one could expect the model to predict $u_n=n/3$, which would be a decent approximation. Yet, our model goes much further, and outputs $u_n=-\cos(3)n/3$, which is a better approximation. We give a few other spectacular examples in Table~\ref{tab:approx-constants}. Our model is remarkably capable of using the values of operators such as $\exp$ and $\arctan$, as if it were able to perform computations internally. 

To investigate the approximation capabilities of our model systematically, we evaluate the its performance when sampling the prefactors uniformly in $[-10,10]$, rather than in $\{-10, -9, ... 9 , 10\} \cup \{e,\pi,\gamma\}$ as done usually. It is impossible for the symbolic model to perfectly represent the correct formulas, but since we are interested in its approximation capabilities, we set the tolerance to $0.01$. Results are shown in Table~\ref{tab:constants-perfs}. Unsurprisingly, the performance of the numeric model is unaffected as it does not suffer from any out-of-vocabulary issues, and becomes better than the symbolic model. However, the symbolic model maintains very decent performances, with its approximation accuracy dropping from 60\% to 42\%. 

This approximation ability in itself an impressive feat, as the model was not explicitly trained to achieve it. It can potentially have strong applications for mathematics, exploited by the recently proposed Ramanujan Machine~\cite{raayoni2021generating}: for example, if a sequence converges to a numerical value such as $1.64493$, it can be useful to ask the model to approximate it, yielding $\pi^2/6$. In fact, one could further improve this approximation ability by training the model only on degree-0 sequences with constant leaves ; we leave this for future work.

\paragraph{Out-of-vocabulary functions}

A similar difficulty arises when dealing with out-of-vocabulary operators, yet again, our model is able to express or approximate them with its own vocabulary.  We show this by evaluating our model on various families of functions from $\texttt{scipy.special}$:
\begin{itemize}[noitemsep]
    \item \textbf{polynomials}: base + orthogonal polynomials of degree 1 to 5 (Legendre, Chebyshev, Jacobi, Laguerre, Hermite, Gegenbauer)
    \item \textbf{hyperbolic}: base + \{sinh, cosh, tanh, arccosh, arcsinh, arctanh\}
    \item \textbf{bessel}: base + \{Bessel and modified Bessel of first and second kinds\}
    \item \textbf{fresnel}: base + \{erf, Faddeeva, Dawson and Fresnel integrals\}.
\end{itemize}

The results in Figure~\ref{fig:operator-families}b show that both the numeric and symbolic models cope surprisingly well with these functions. The symbolic model has more contrasted results than the numeric model, and excels particularly on functions which it can easily build with its own vocabulary such as polynomials. Surprisingly however, it also outperforms the numeric model on the other groups.

In Table~\ref{tab:approx-functions}, we show a few examples of the remarkable ability of our model to yield high-quality asymptotic approximations, either using a recurrence relation as for the Catalan numbers and the Dawson function, either with a non-recurrent expression as for the Bessel functions.

\section{Conclusion}

In this work, we have shown that Transformer models can successfully infer recurrence relations from observations. We applied our model to challenging out-of-distribution tasks, showing that it outperforms Mathematica functions on integer sequences and yields informative approximations of complex functions as well as numerical constants. 

\paragraph{Scope of our approach}
One may ask to what extent our model can be used for real-world applications, such as time-series forecasting. Although robustness to noise is an encouraging step in this direction, we believe our model is not directly adapted to such applications, for two reasons.

First, real-world time-series often cannot be described by a simple mathematical formula, in which case numeric approaches will generally outperform symbolic approaches. Second, even when they can be expressed by a formula, the latter will contain complex prefactors and non-deterministic terms which will make the task extremely challenging.

\paragraph{Future directions}
To bring our model closer to real-world problems, we need our model to be able to handle arbitrary prefactors. We introduce a method to solve this problem in our follow-up work by introducing numeric tokens in the decoder~\cite{kamienny2022end}.

Another important extension of our work is the setting of multi-dimensional sequences. With two dimensions, one could study complex sequences, which are a well-studied branch of mathematics given their relationship with fractals. 

Finally, recurrence relations being a discrete version of differential equations, the most natural extension to this work is to infer differential equations from trajectories; this will be an important direction for future work.




%% file: appendix.tex
\onecolumn

\section{Robustness to noise}
\label{app:stochastic}

One particularity of our model is that it is entirely trained and evaluated on synthetic data which is completely noise-free. Can our model also predict recurrence relations when the inputs are corrupted ? In this section, we show that the answer is yes, provided the model is trained with noisy inputs. For simplicity, we restrict ourselves here to the setup of float sequences, but the setup of integer sequences can be dealt with in a similar manner, trading continuous random variables for discrete ones.

\paragraph{Setup}

Considering the wide range of values that are observed in recurrent sequences, corruption via additive noise with constant variance, i.e. $u_n = f(n, \{u_i\}_{i<n}) + \xi_n, \xi_n\sim\mathcal{N}(0,\sigma)$ is a poor model of stochasticity. Indeed, the noise will become totally negligible when $u_n\gg 1$, and conversely, totally dominate when $u_n\ll 1$. To circumvent this, we scale the variance of the noise with the magnitude of the sequence, i.e. $\xi_n\sim\mathcal{N}(0,\sigma u_n)$, allowing to define a signal-to-noise ratio $\mathrm{SNR}=1/\sigma$. This can also be viewed as a multiplicative noise $u_n = f(n, \{u_i\}_{i<n})\xi, \xi\sim\mathcal{N}(1,\sigma)$.

\paragraph{Results}

To make our models robust to corruption in the sequences, we use stochastic training. This involves picking a maximal noise level $\sigma_{train}$, then for each input sequence encountered during training, sample $\sigma\sim \mathcal{U}(0,\sigma_{train})$, and corrupt the terms with a multiplicative noise of variance $\sigma$. At test time, we corrupt the input sequences with a noise of fixed variance $\sigma_{test}$, but remove the stochasticity for next term prediction, to check whether our model correctly inferred the deterministic part of the formula.

Results are presented in Table~\ref{tab:noise}. We see that without the stochastic training, the accuracy of our model plummets from 43\% to 1\% as soon as noise is injected at test time. However, with stochastic training, we are able to keep decent performance even at very strong noise levels: at $\sigma_{test}=0.5$, we are able to achieve an accuracy of 17\%, which is remarkable given that the signal-to-noise ratio is only of two. However, this robustness comes at a cost: performance on the clean dataset is degraded, falling down to 30\%.

\begin{table}[h]
    \centering
    \begin{tabular}{c|c|c|c}
    \toprule
        $\sigma_{train/test}$ & $\sigma_{test}=0$ & $\sigma_{test}=0.1$ & $\sigma_{test}=0.5$\\
        \midrule
        $\sigma_{train}=0$      & \textbf{43.3} & 0.9 & 0.0 \\
        $\sigma_{train}=0.1$    & 38.4 & \textbf{31.9} & 0.2 \\        
        $\sigma_{train}=0.5$    & 35.6 & 31.8 & \textbf{11.1} \\
    \bottomrule
    \end{tabular}
    \caption{\textbf{Our symbolic model can be made robust to noise in the inputs, with a moderate drop in performance on clean inputs.} We report the accuracy on expressions with up to 10 operators, for $n_{pred}=10$, $\tau=10^{-10}$, varying the noise level during training $\sigma_{train}$ and evaluation $\sigma_{test}$.}
    \label{tab:noise}
\end{table}

\section{Robustness to distribution shifts}
\label{app:ood-init}

Note that in all our training examples, the initial terms of the sequences are sampled in the range $[-10,10]$. This naturally begs the question: can our model handle cases where the first terms lie outside this range? 

Results, shown in Tab.~\ref{tab:ood-init}, show that our symbolic models displays poor robustness to this distribution shift: their accuracy plummets by more than a factor of two. In comparison, the numeric model is significantly less affected.

Note that an easy fix to this issue would be to train with a larger range for the scale of the initial terms; nonetheless, this confirms the lack of out-of-distribution robustness of numeric-to-symbolic models, previously demonstrated by~\cite{charton2021linear}.

\begin{table}[htb]
    \centering
    \small
    \begin{tabular}{c|cc|cc}
    \toprule
        \multirow{2}{*}{Model} & \multicolumn{2}{c|}{Integer} & \multicolumn{2}{c}{Float}\\
        & $[\![-10, 10]\!]$ & $[\![-100, 100]\!]$ & $[-10, 10]$ & $[-100, 100]$ \\
        \midrule
        Symbolic & 80.3 & 30.5 & 60.7 & 27.9\\
        Numeric  & 75.5 & 67.4 & 60.4 & 60.2\\
    \bottomrule
    \end{tabular}
    \caption{\textbf{The symbolic models suffer from distribution shifts.} We report the in-domain accuracies obtained when sampling the first terms of the sequences uniformly in $[\![-100, 100]\!]$ (integer) and $[-100, 100]$ (float) instead of $[\![-10, 10]\!]$ and $[-10, 10]$ seen during training. We set $\tau=0.01$ and $n_{pred}=10$.}
    \label{tab:ood-init}
\end{table}

\section{The effect of expression simplification}
\label{app:sympy}

One issue with symbolic regression is the fact that a mathematical expression such as $\texttt{mul, 2, cos, n}$ can be written in many different ways. Hence, cross-entropy supervision to the tokens of the expression can potentially penalize the model for generating the same formula written in a different way (e.g. $\texttt{mul, cos, n, 2}$ or $\texttt{mul, cos, n, add, 1, 1}$). To circumvent this issue,~\cite{petersen2019deep} first predict the formula, then evaluate it and supervise the evaluations to those of the target function. Yet, since the evaluation step is non-differentiable, they are forced to use a Reinforcement Learning loop to provide reward signals. In our framework, we noticed that such an approach is actually unnecessary.

\begin{figure}[htb]
    \centering
    \begin{subfigure}[b]{.24\textwidth}
    \includegraphics[width=\linewidth]{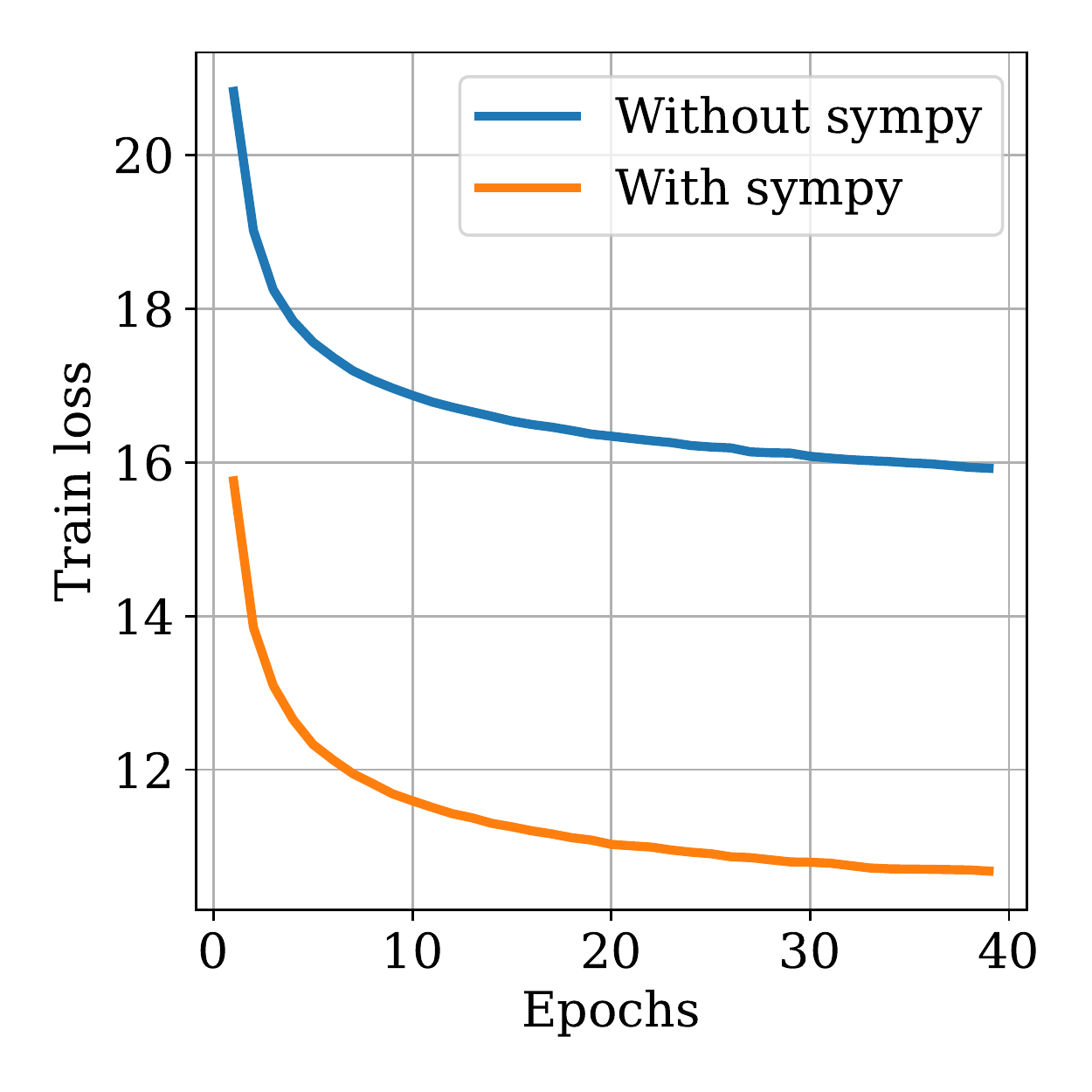}
    \caption{Training loss, integer}
    \end{subfigure}
    \begin{subfigure}[b]{.24\textwidth}
    \includegraphics[width=\linewidth]{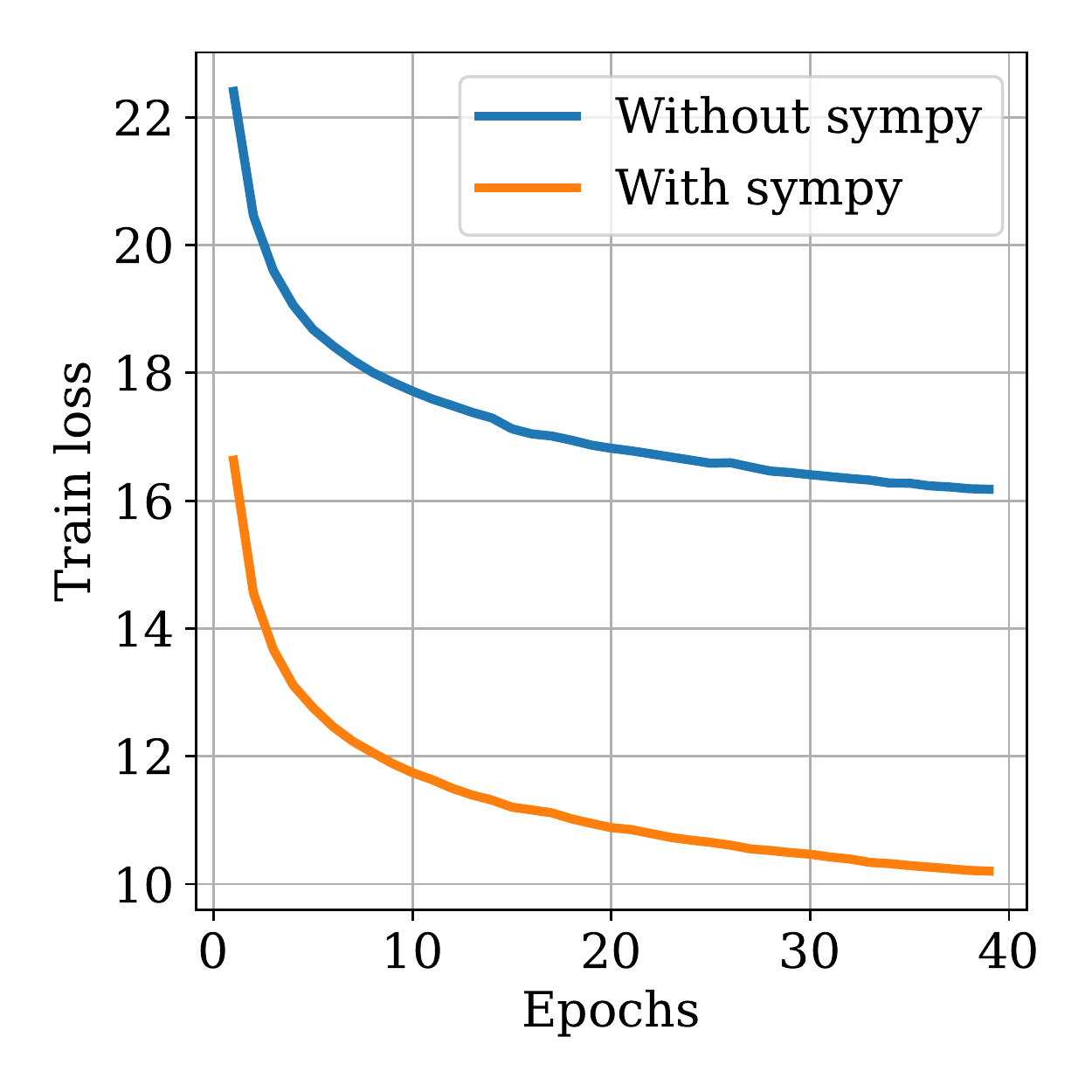}
    \caption{Training loss, float}
    \end{subfigure}
    \begin{subfigure}[b]{.24\textwidth}
    \includegraphics[width=\linewidth]{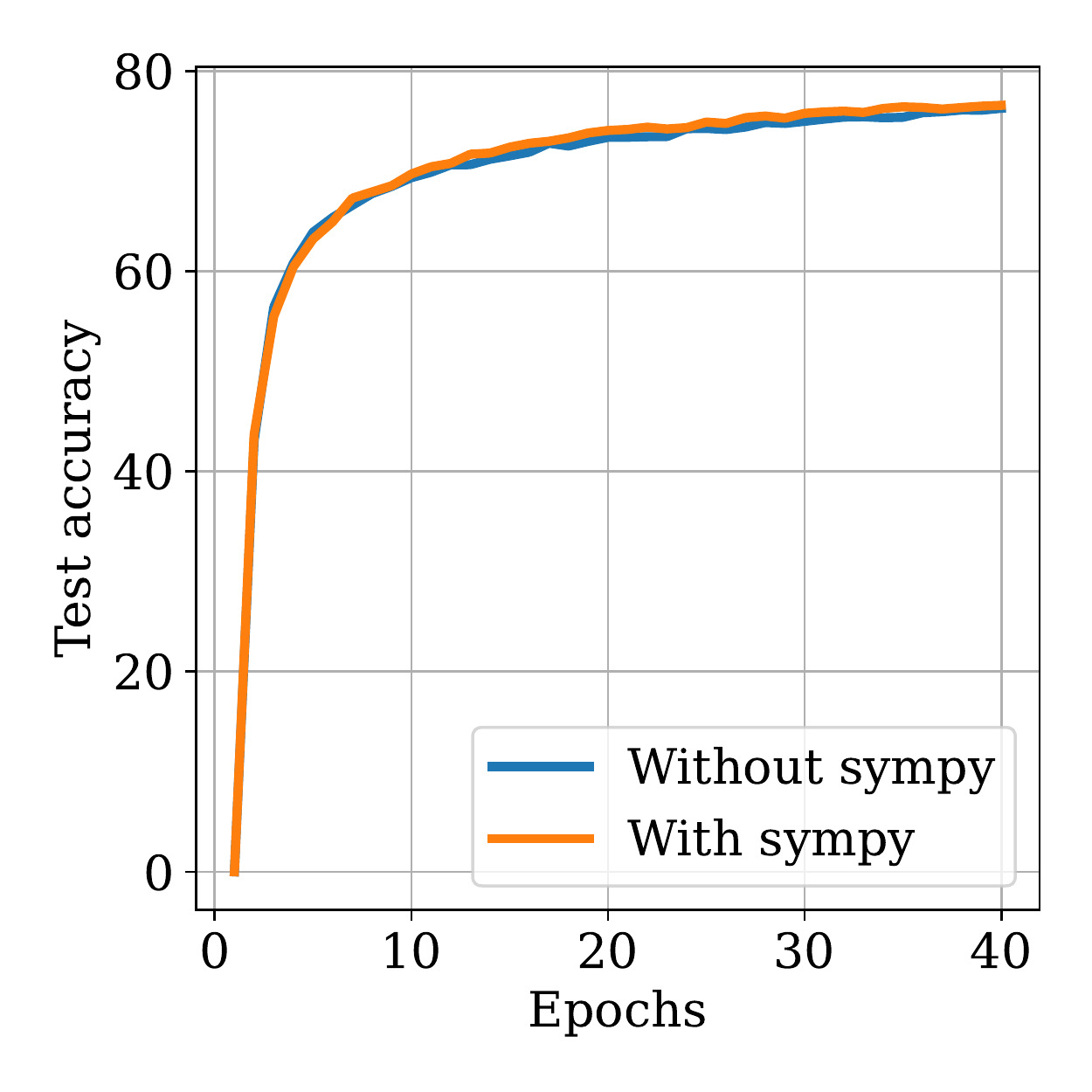}
    \caption{Test accuracy, integer}
    \end{subfigure}
    \begin{subfigure}[b]{.24\textwidth}
    \includegraphics[width=\linewidth]{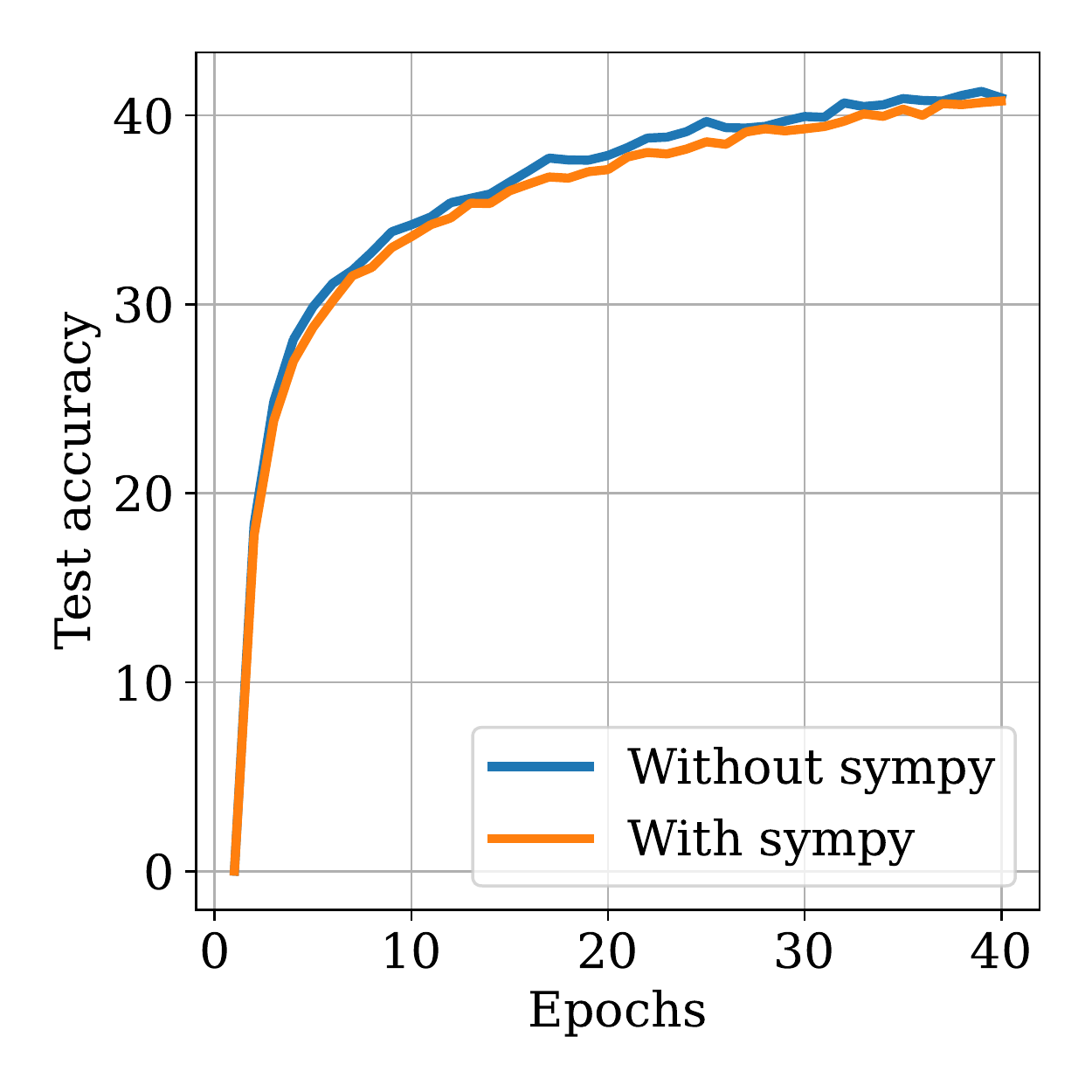}
    \caption{Test accuracy, float}
    \end{subfigure}
    \caption{\textbf{Simplification reduces the training loss, but does not bring any improvement in test accuracy}. We displayed the first 40 epochs of training of our symbolic models.}
    \label{fig:sympy}
\end{figure}

Instead, one could simply preprocess the mathematical formula to simplify it with SymPy~\cite{meurer2017sympy} before feeding it to the model. This not only simplifies redundant parts such as $\texttt{add, 1, 1} \to \texttt{2}$, but also gets rid of the permutation invariance $\texttt{mul, x, 2} = \texttt{mul, 2, x}$ by following deterministic rules for the order of the expressions. However, and rather surprisingly, we noticed that this simplification does not bring any benefit to the predictive power of our model : although it lowers the training loss (by getting rid of permutation invariance, it lowers cross-entropy), it does not improve the test accuracy, as shown in Fig.~\ref{fig:sympy}. This suggests that expression syntax is not an issue for our model: the hard part of the problem indeed lies in the mathematics.

Aside from predictive power, SymPy comes with several advantages and drawbacks. On the plus side, it enables the generated expressions to be written in a cleaner way, and improves the diversity of the beam. On the negative side, it slows down training, both because it is slow to parse complex expressions, and because it often lengthens expression since it does not handle division (SymPy rewrites $\texttt{div,a,b}$ as $\texttt{mul,a,pow,b,-1}$). Additionally, simplification actually turns out to be detrimental to the out-of-domain generalization of the float model. Indeed, to generate approximations of out-of-vocabulary prefactors, the latter benefits from non-simplified numerical expressions. Hence, we chose to not use SymPy in our experiments.

\section{Does memorization occur?}
\label{app:memorization}

It is natural to ask the following question: due to the large amount of data seen during training, is our model simply memorizing the training set ? Answering this question involves computing the number of possible inputs sequences $N_{seq}$ which can be generated. To estimate this number, calculating the number of possible mathematical expressions $N_{expr}$ is insufficient, since a given expression can give very different sequences depending on the random sampling of the initial terms. Hence, one can expect that $N_{expr}$ is only a very loose lower bound for $N_{seq}$. 

\begin{figure}[h]
    \centering
    \includegraphics[width=.3\columnwidth]{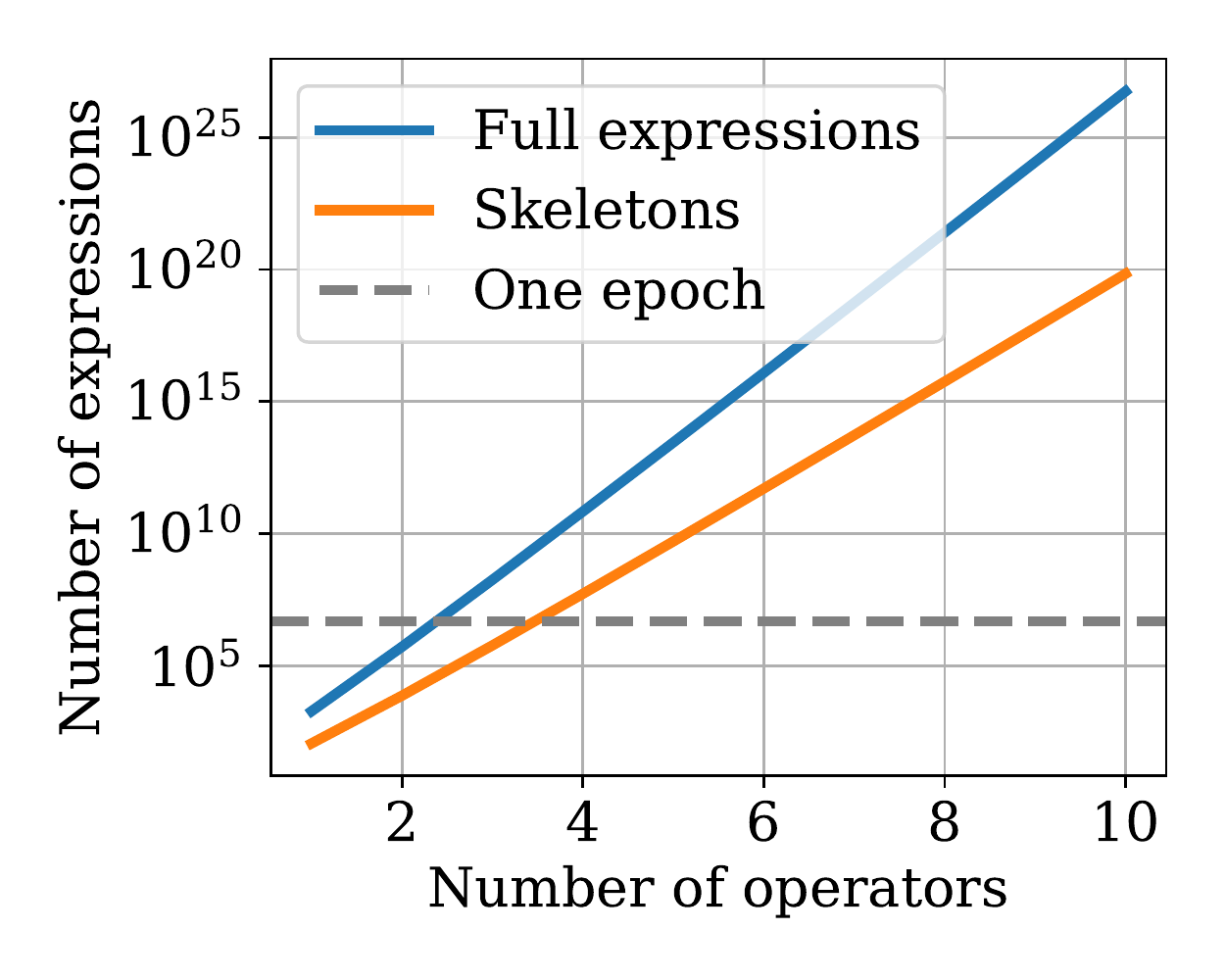}
    \caption{\textbf{Our models only see a small fraction of the possible expressions during training.} We report the number of possible expressions for each number of operators (skeleton refers to an expression with the choice of leaves factored out). Even after a hundred epochs, our models have only seen a fraction of the possible expressions with more than 4 operators.}
    \label{fig:num_expressions}
\end{figure}

Nonetheless, we provide the lower bound $N_{expr}$ as a function of the number of nodes in Fig.~\ref{fig:num_expressions}, using the equations provided in~\cite{lample2019deep}. For small expressions (up to four operators), the number of possible expressions is lower or similar to than the number of expressions encountered during training, hence one cannot exclude the possibility that some expressions were seen several times during training, but with different realizations due to the initial conditions. However, for larger expressions, the number of possibilities is much larger, and one can safely assume that the expressions encountered at test time have not been seen during training.

\section{The issue of finite precision}
\label{app:iterative-refinement}

As explained in the main text, our model tends to ignore subdominant terms in expressions with terms of vastly different magnitudes, partly due to finite precision. For example, when using a float precision of $p=4$ digits, we obtain a discretization error of $10^5$ for numbers of magnitude $10^9$. However, there exists two methods to circumvent this issue.

\paragraph{Increasing the precision}

Naively increase the precision $p$ would cause the vocabulary size of the encoder to rapidly explode, as it scales as $10^p$. However, one can instead encode the mantissa on multiple tokens, as performed for integers. For example, using two tokens instead of one, e.g. encoding $\pi$ as $\texttt{+ 3141, 5926, E-11}$, doubles the precision while increasing the sequence length only by 33\%. 

This approach works well for recurrence prediction, but slightly hampers the ability of the model to approximate prefactors as shown in Tab.~\ref{tab:approx-constants}, hence we did not use it in the runs presented in this paper.

\paragraph{Iterative refinement}

Another method to improve the precision of the model is to use an iterative refinement of the predicted expression, akin to perturbation theory in physics.

Consider, for example, the polynomial $f(x)=\sum_{k=0}^d a_k x^k$, for which our model generally predicts $\hat f(x) = \sum_{k=0}^d \hat a_k x^k$, with the first coefficient correct ($\hat a_d=a_d$) but potentially the next coefficients incorrect ($\hat a_k \neq a_k$ for $k<d$). One can correct these subdominant coefficients iteratively, order by order. To obtain the term $a_{d-1}$, fit the values of $g(x) = f(x)-\hat f(x) = \sum_{k=1}^{d-1} (a_k - \hat a_k)$. Then fit the values of $h(x) = g(x)-\hat g(x)$, etc. By iterating this procedure $k$ times, one can obtain the $k$ highest coefficients $a_k$. 

We checked that this method allows us to approximate any polynomial function. One could in fact use iterative refinement to predict the Taylor approximation of any function, or use a similar approach to catch multiplicative corrections, by fitting $g(x) = f(x)/\hat f(x)$; we leave these investigations for future work.

\begin{figure*}[h]
    \centering
    \begin{subfigure}[b]{.24\textwidth}
    \includegraphics[width=\linewidth]{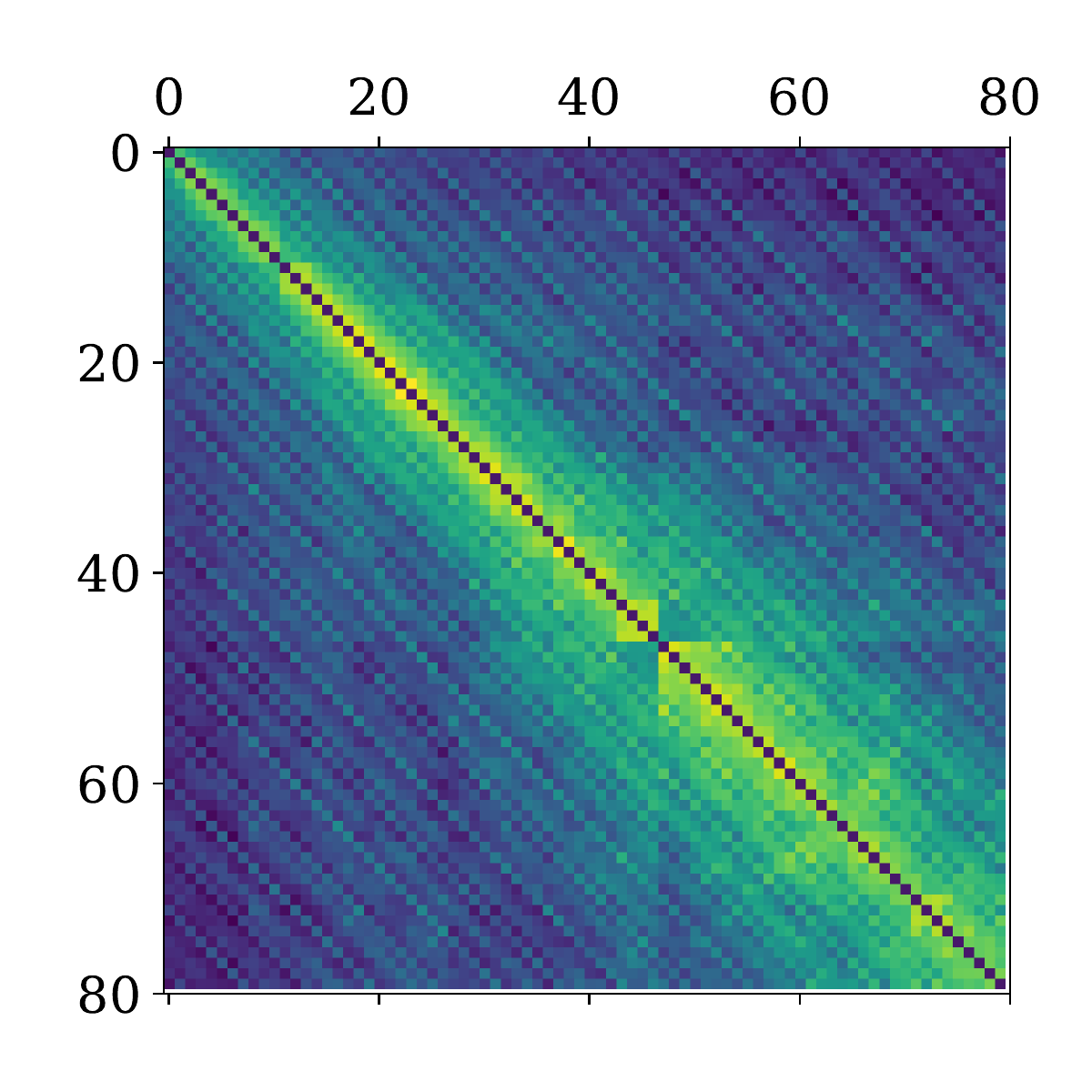}
    \caption{Integer symbolic}
    \end{subfigure} 
    \begin{subfigure}[b]{.24\textwidth}
    \includegraphics[width=\linewidth]{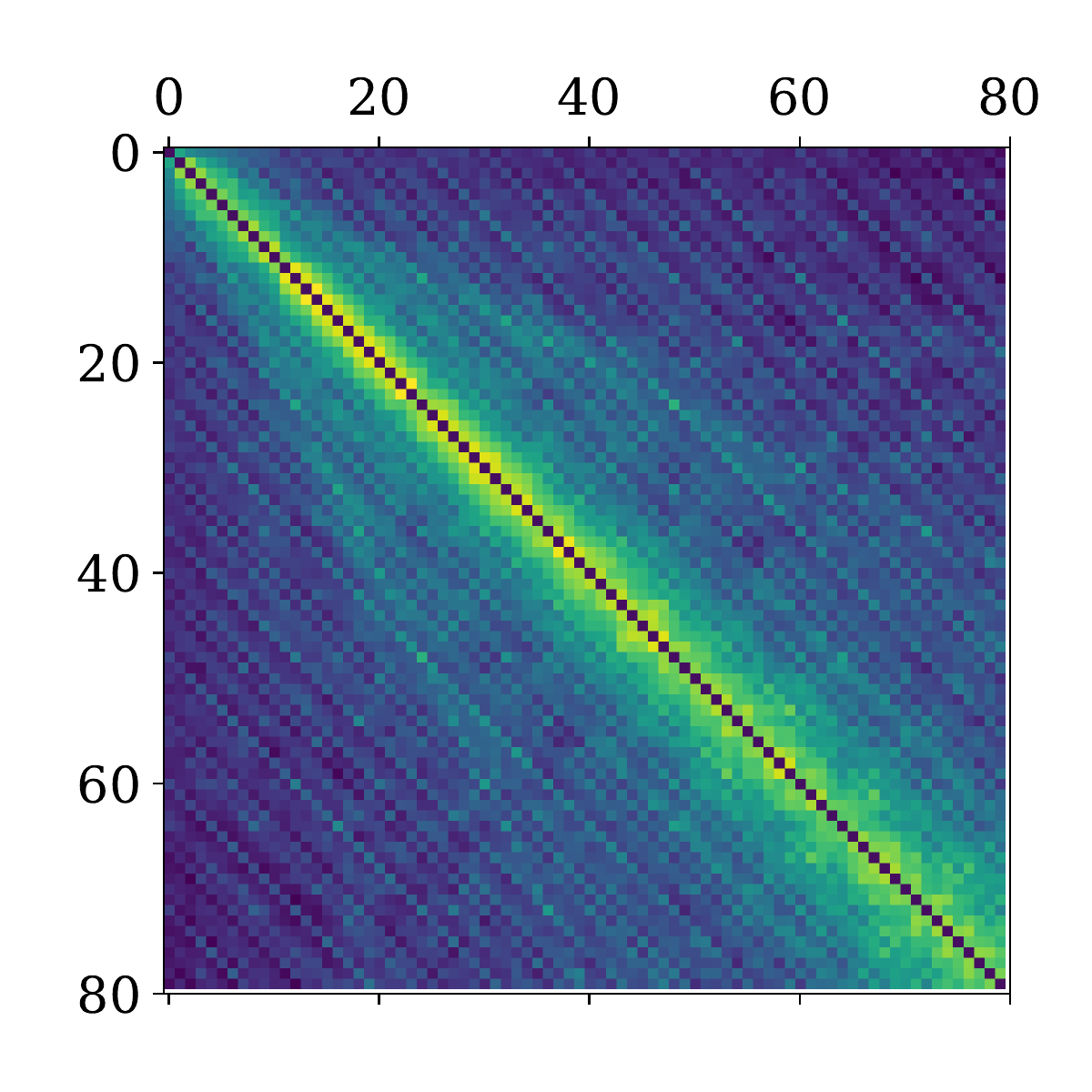}
    \caption{Integer numeric}
    \end{subfigure}
    \begin{subfigure}[b]{.24\textwidth}
    \includegraphics[width=\linewidth]{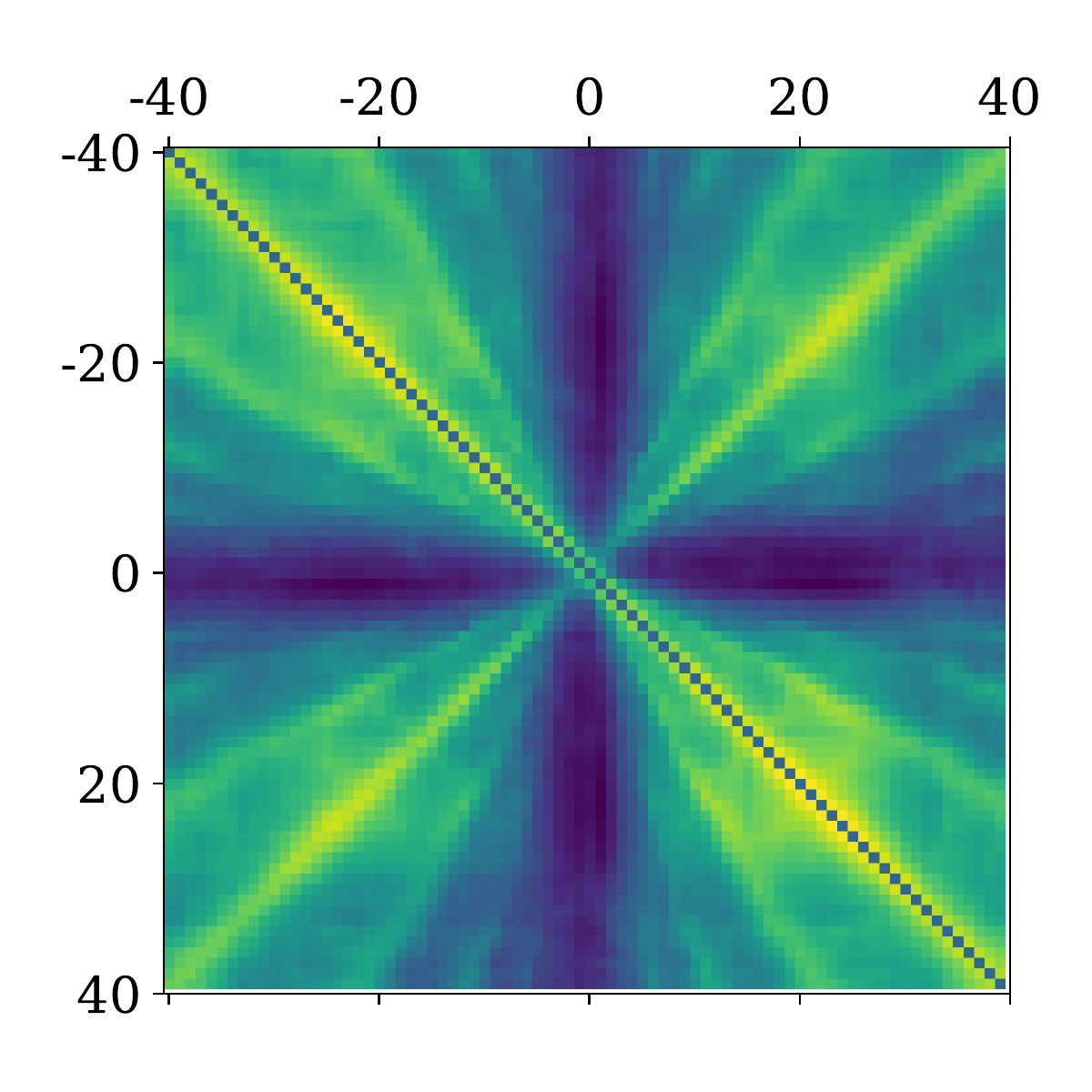}
    \caption{Float symbolic}
    \end{subfigure}
    \begin{subfigure}[b]{.24\textwidth}
    \includegraphics[width=\linewidth]{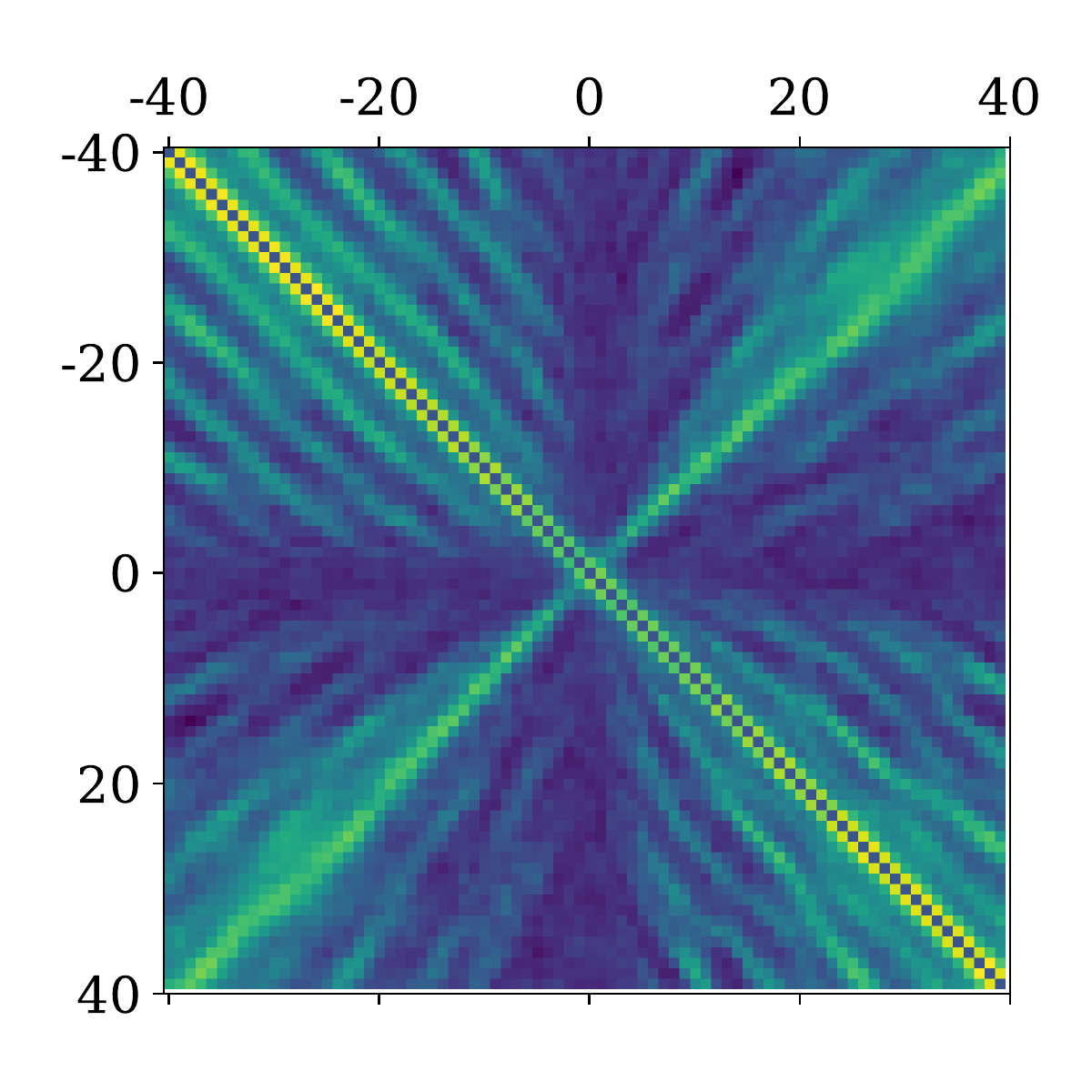}
    \caption{Float numeric}
    \end{subfigure}
    \caption{\textbf{The similarity matrices reveal more details on the structure of the embeddings}. The element $(i,j)$ is the cosine similarity between embeddings $i$ and $j$.}
    \label{fig:embeddings-similarity}
\end{figure*}

\section{Structure of the embeddings}
\label{app:embeddings}
To gain better understanding on the number embeddings of our models, we depict similarity matrices whose element $(i,j)$ is the cosine similarity between embeddings $i$ and $j$ in Fig.~\ref{fig:embeddings-similarity}. 

\paragraph{Integer sequences}
The numeric and symbolic similarity matrices look rather similar, with a bright region appearing around the line $y=x$, reflecting the sequential nature of the embeddings. In both cases, we see diagonal lines appear : these correspond to lines of common divisors between integers. Strikingly, these lines appear most clearly along multiples of 6 and 12, especially in the symbolic model, suggesting that 6 is a natural base for reasoning. These results are reminiscent of the much earlier explorations of ~\citet{paccanaro2001learning}.

\paragraph{Float sequences}
Both for the numeric and symbolic setups, the brightest regions appear along the diagonal lines $y=x$ and $y=-x$, reflecting respectively the sequential nature of the embeddings and their symmetry around 0. The darkest regions appears around the vertical line $x=0$ and the horizontal line $y=0$, corresponding to exponents close to zero: these exponents strongly overlap with each other, but weakly overlap with the rest of the exponents. Interestingly, dimmer lines appear in both setups, but follow a very different structure. In the symbolic setup, the lines appear along lines $y=\pm x/k$, reminiscent of the effect of polynomials of degree $k$. In the numeric setup, the lines are more numerous, all diagonal but offset vertically.



\section{Visualizations}
\label{app:examples}

\paragraph{Success and failure modes}

\begin{figure*}
    \centering
    \begin{subfigure}[b]{.49\textwidth}
    \includegraphics[width=\linewidth]{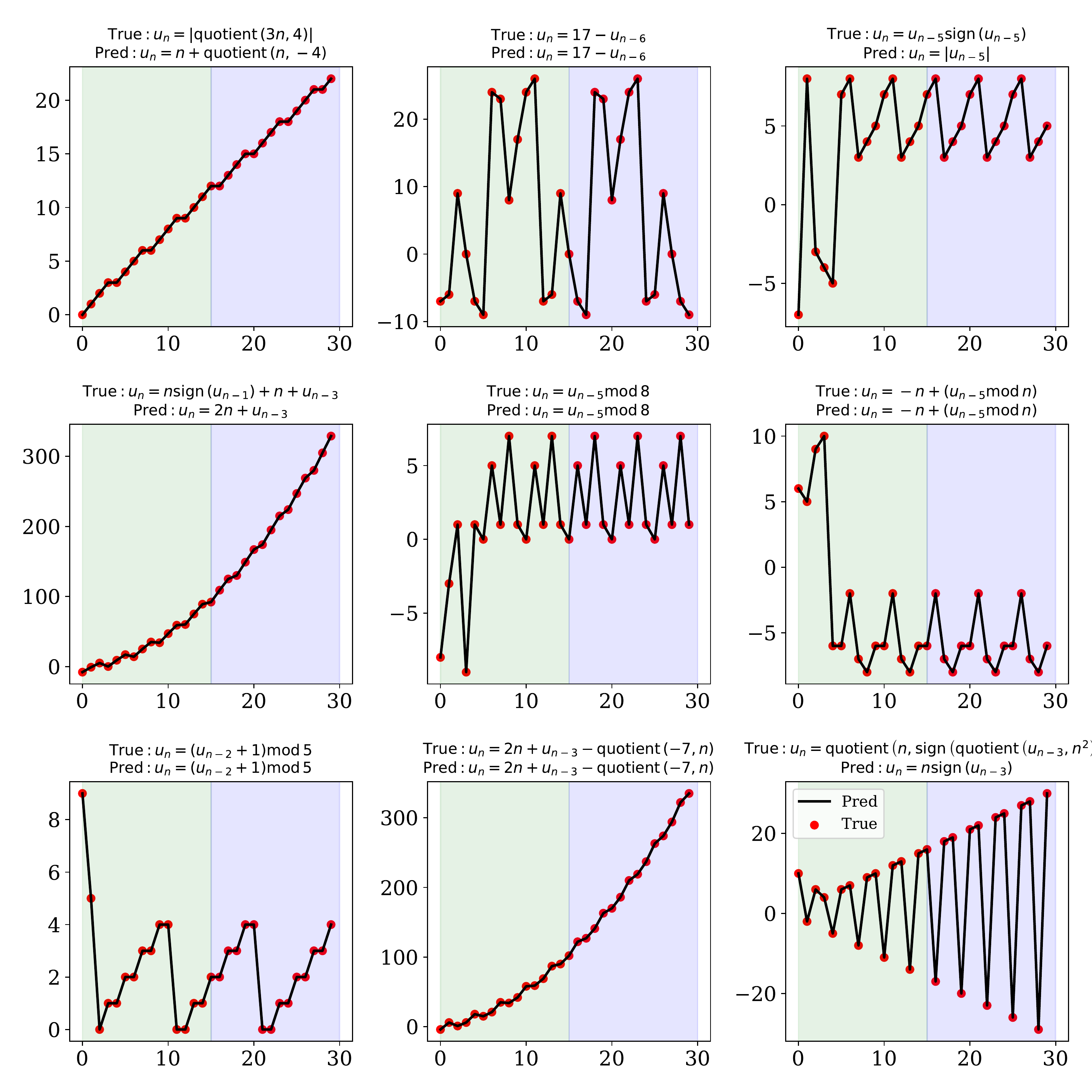}
    \caption{Integer, success}
    \end{subfigure} 
    \begin{subfigure}[b]{.49\textwidth}
    \includegraphics[width=\linewidth]{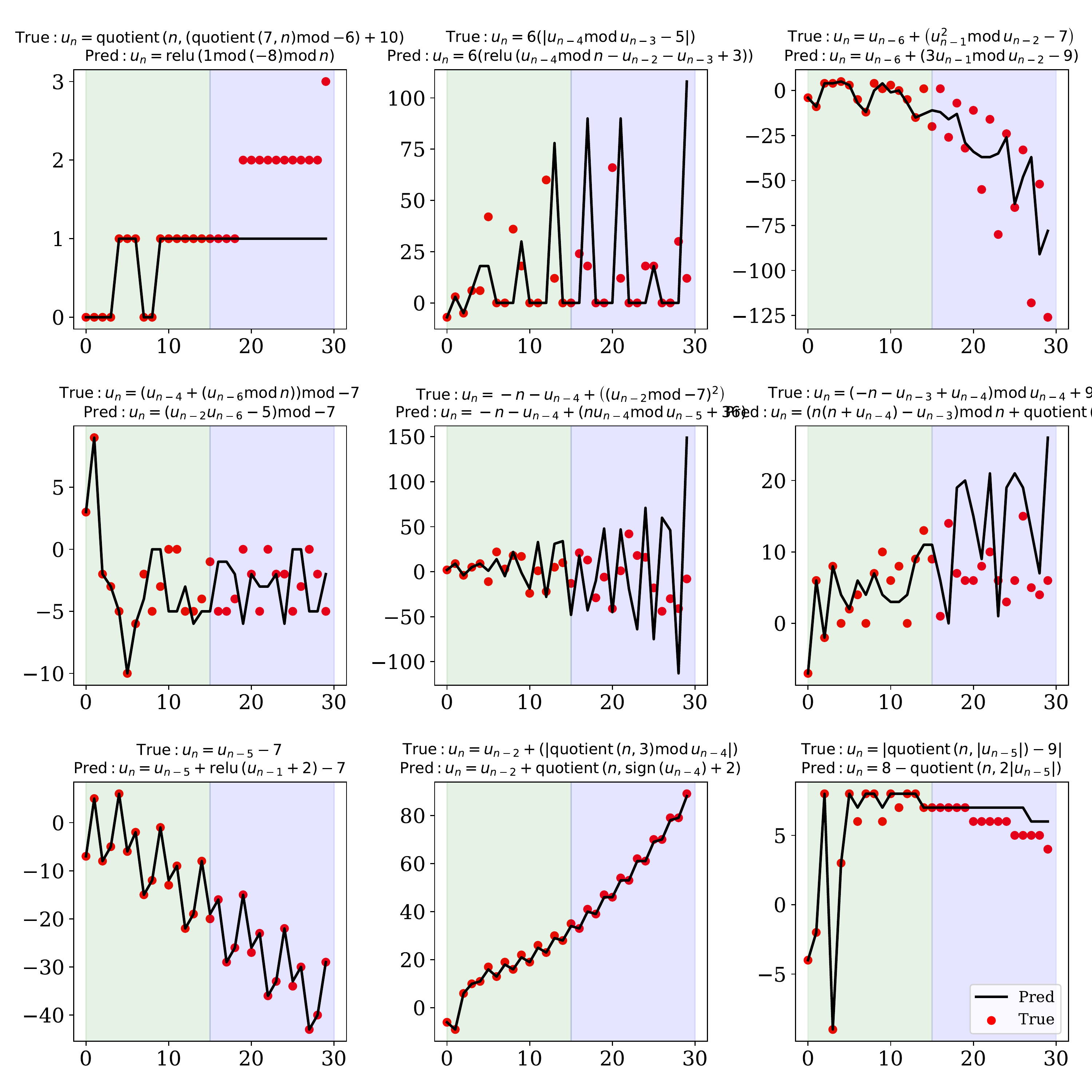}
    \caption{Integer, failure}
    \end{subfigure} 
    \begin{subfigure}[b]{.49\textwidth}
    \includegraphics[width=\linewidth]{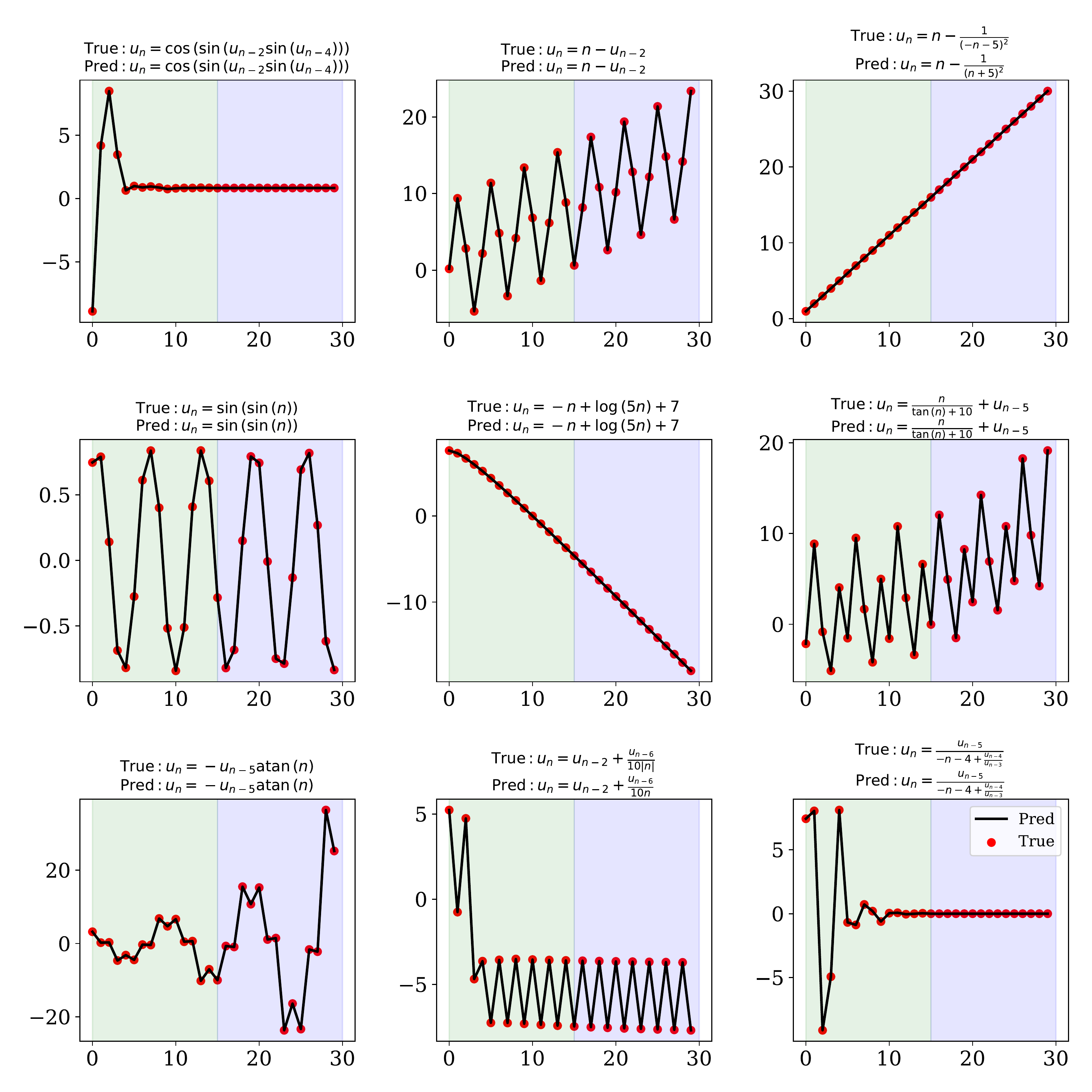}
    \caption{Float, success}
    \end{subfigure} 
    \begin{subfigure}[b]{.49\textwidth}
    \includegraphics[width=\linewidth]{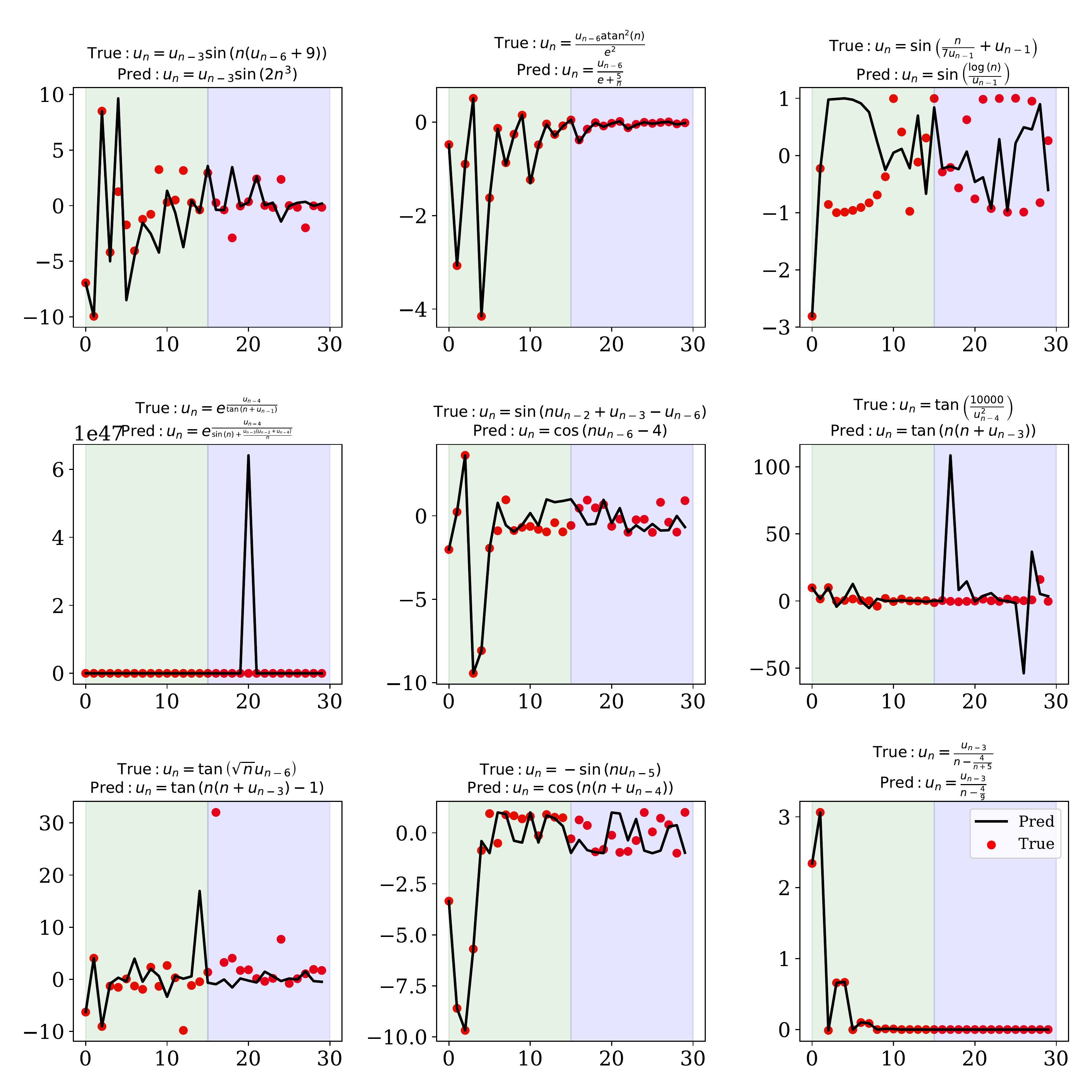}
    \caption{Float, failure}
    \end{subfigure} 
    \caption{\textbf{Success and failure modes of our models.} The models are fed the first 15 terms of the sequence (green area) and predict the next 15 terms (blue area). We randomly selected expressions with 4 operators from our generator, and picked the first successes and failures.}
    \label{fig:success-failure}
\end{figure*}

In Fig.~\ref{fig:success-failure}, we show a few examples of success and failure modes of our symbolic models. The failure modes are particularly interesting, as they reflect the strong behavioral difference between our symbolic models and models usually used for regression tasks.

The latter generally try to interpolate the values of the function they are given, whereas our symbolic model tries to predict the expression of the function. Hence, our model cannot simply "overfit" the inputs. A striking consequence of this is that in case of failure, the predicted expression is wrong both on the input points (green area) and the extrapolation points (blue area).

In some cases, the incorrectly predicted formula provides a decent approximation of the true function (e.g. when the model gets a prefactor wrong). In others, the predicted formula is catastrophically wrong (e.g. when the model makes a mistake on an operator or a leaf).

\paragraph{Training curves}
\label{app:training-curves}

\begin{figure*}
    \centering
    \begin{subfigure}[b]{\linewidth}
    \includegraphics[width=\columnwidth]{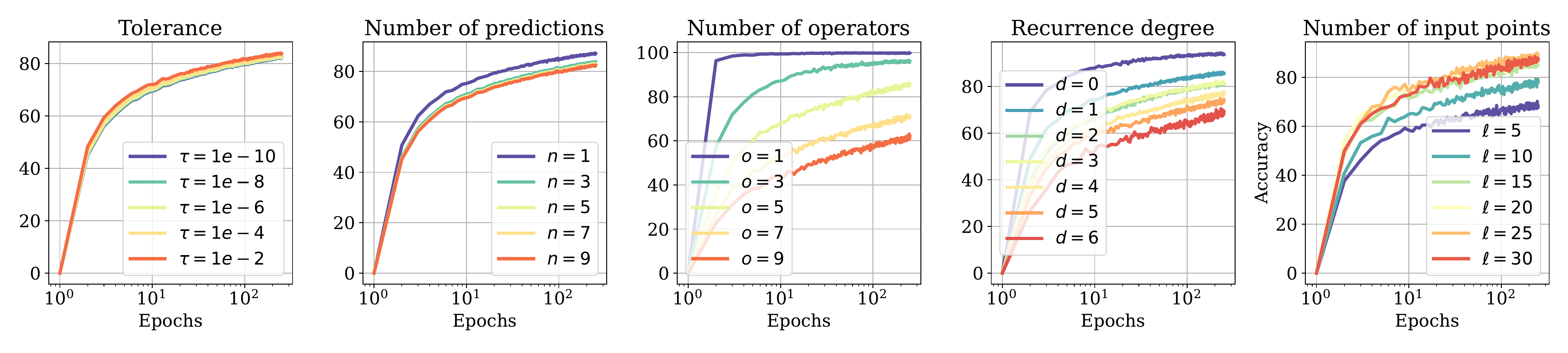}
    \caption{Integer symbolic}
    \end{subfigure} 
    \begin{subfigure}[b]{\linewidth}
    \includegraphics[width=\linewidth]{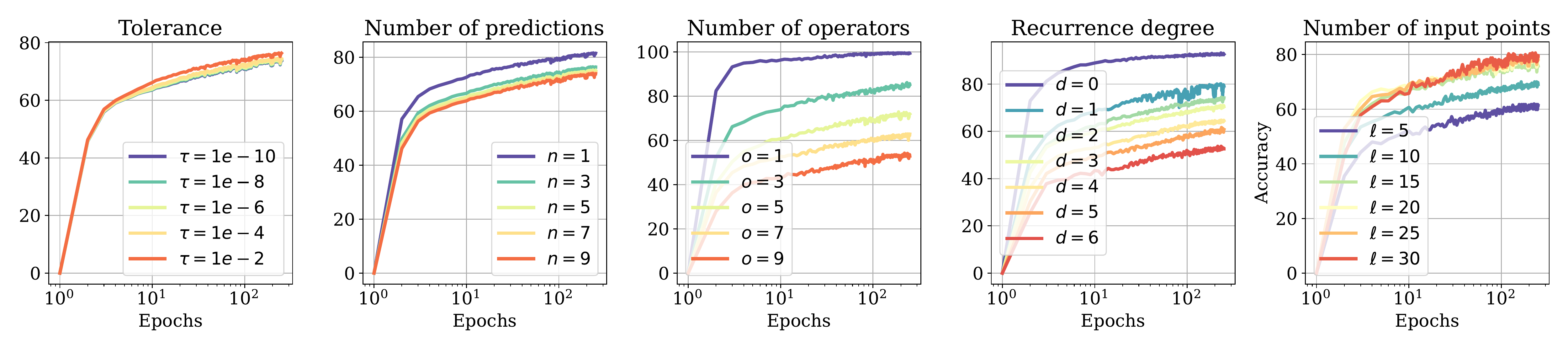}
    \caption{Integer numeric}
    \end{subfigure} 
    \begin{subfigure}[b]{\linewidth}
    \includegraphics[width=\linewidth]{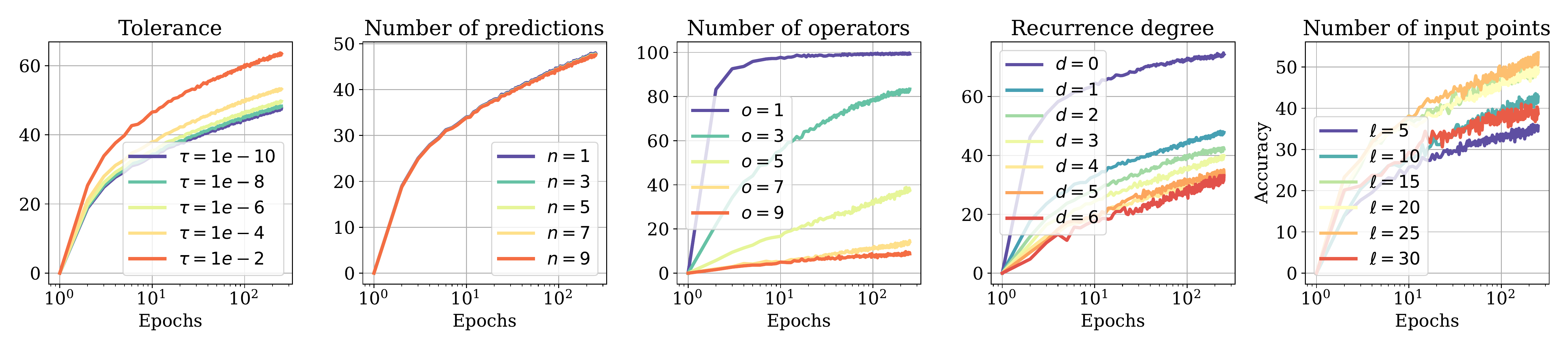}
    \caption{Float symbolic}
    \end{subfigure} 
    \begin{subfigure}[b]{\linewidth}
    \includegraphics[width=\linewidth]{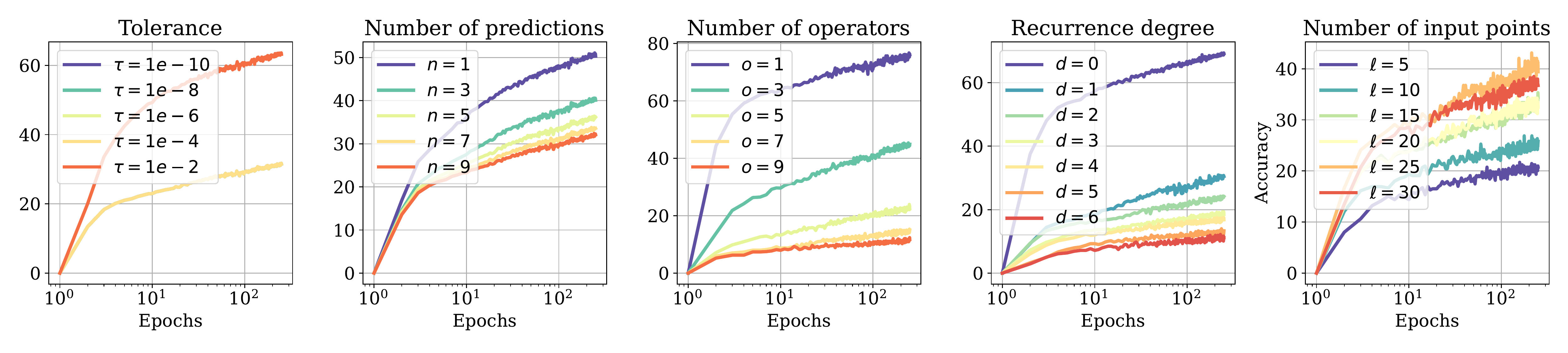}
    \caption{Float numeric}
    \end{subfigure} 
    \caption{\textbf{Training curves of our models.} We plot the accuracy of our models at every epoch, evaluated of 10,000 sequences generated from the same distribution as during trianing. From left to right, we vary the tolerance $\tau$, the number of predictions $n_{pred}$, the number of operators $o$, the recurrence degree $d$ and the number of input terms $l$. In each plot, we use the following defaults for quantities which are not varied: $\tau=10^{-10}$, $n_{pred}=10$, $o\in[\![1,10]\!]$, $d\in[\![1,6]\!]$, $l\in[\![5,30]\!]$.}
    \label{fig:time}
\end{figure*}

In Fig.~\ref{fig:time}, we show the training curves of our models, presenting an ablation over the tolerance $\tau$, the number of predictions $n_{pred}$, the number of operators $o$, the recurrence degree $d$ and the number of input points $l$, as explained in the main text.

\paragraph{Attention maps}
\label{app:attention}

\begin{figure*}
    \centering
    \begin{subfigure}[b]{\textwidth}
    \includegraphics[width=\linewidth]{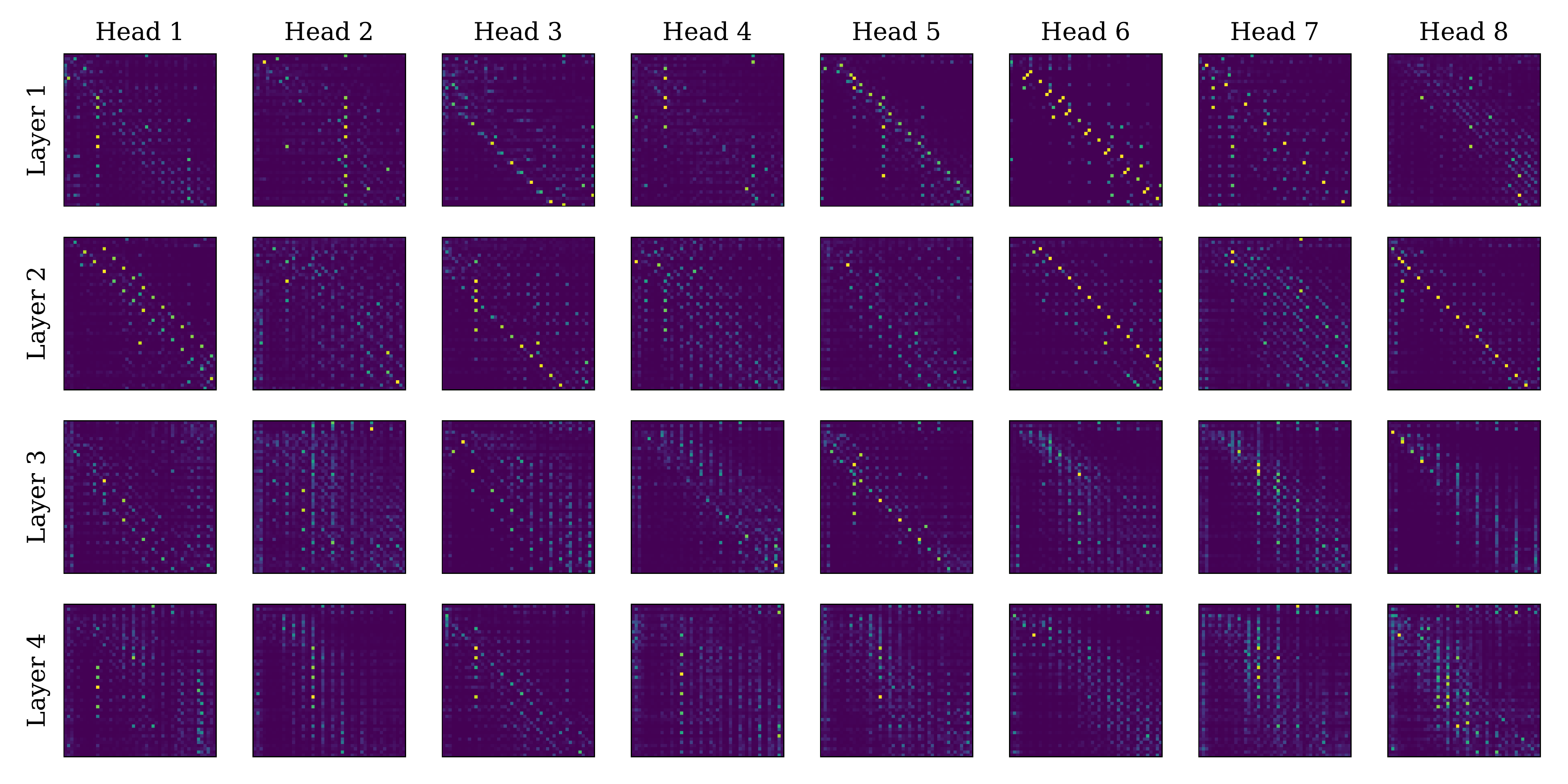}
    \caption{Integer}
    \end{subfigure} 
    \begin{subfigure}[b]{\textwidth}
    \includegraphics[width=\linewidth]{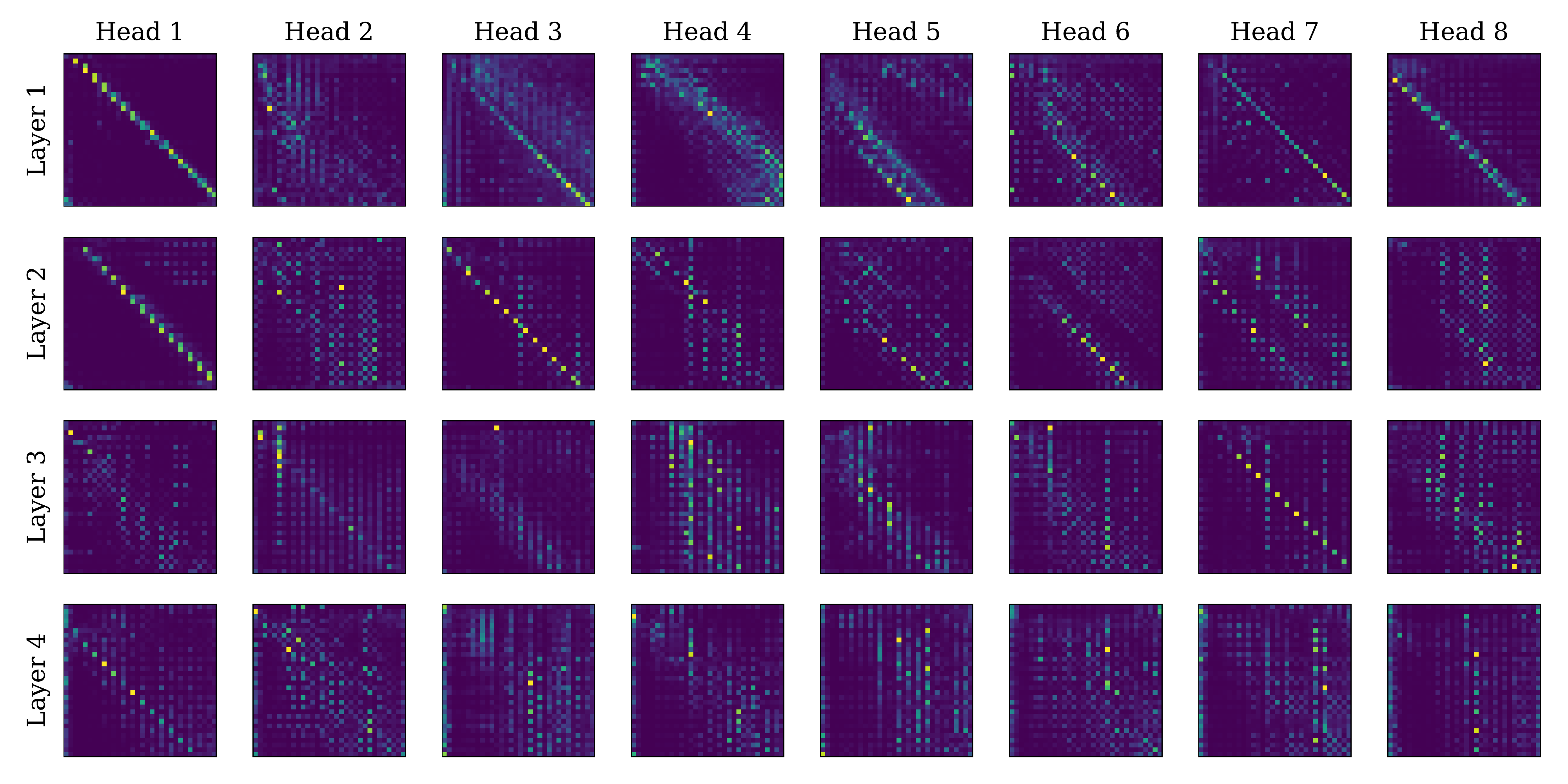}
    \caption{Float}
    \end{subfigure} 
    \caption{\textbf{Attention maps of our integer model and float models}. We evaluated the integer model on the first 25 terms of the sequence $u_n = -(6+u_{n-2})\operatorname{mod} n$ and the float model on the first 25 terms of the sequence $u_n = \exp(\cos(u_{n-2}))$.}
    \label{fig:attention}
\end{figure*}

In Fig.~\ref{fig:attention}, we provide attention maps for the 8 attention heads and 4 layers of our Transformer encoders. Clearly, different heads play very different roles, some focusing on local interactions and others on long-range interactions. However, the role of different layers is hard to interpret.

